%% file: acl_latex.tex
\pdfoutput=1

\documentclass[11pt]{article}

\usepackage{acl}

\usepackage{times}
\usepackage{latexsym}

\usepackage[T1]{fontenc}

\usepackage[utf8]{inputenc}

\usepackage{microtype}

\usepackage{inconsolata}

\usepackage{svg}
\usepackage{inconsolata}
\usepackage{times}
\usepackage{colortbl}
\usepackage{latexsym}
\usepackage{amsmath}
\usepackage{amssymb}
\usepackage{array}
\usepackage{pifont}
\usepackage{microtype}
\usepackage{tabularx}
\usepackage{adjustbox}
\usepackage{enumitem}
\usepackage{multirow}
\usepackage{dsfont}
\usepackage{booktabs}
\usepackage{algorithm2e}
\usepackage{MnSymbol}
\usepackage{scalerel}
\usepackage{mathrsfs}
\usepackage{pifont}
\usepackage{multirow}
\usepackage{graphicx}
\usepackage{changes}
\usepackage{xcolor,soul}
\usepackage{makecell}
\usepackage{ulem}
\usepackage{graphicx,calc}
\usepackage{bbm}
\usepackage{lipsum}
\usepackage[caption=false]{subfig}

\newcommand{\Figref}[1]{Figure~\ref{#1}}
\newcommand{\figref}[1]{Figure~\ref{#1}}
\newcommand{\Tabref}[1]{Table~\ref{#1}}

\newcommand{\Secref}[1]{Section~\ref{#1}}

\newcommand{\Appref}[1]{Appendix~\ref{#1}}

\usepackage[most]{tcolorbox}

\newtcbox{\hlprimarytab}{on line, rounded corners, box align=base, colback=white!10,colframe=white,size=fbox,arc=3pt, before upper=\strut, top=-2pt, bottom=-4pt, left=-2pt, right=-2pt, boxrule=0pt}
\newtcbox{\hlprimarytabg}{on line, rounded corners, box align=base, colback=gray!10,colframe=white,size=fbox,arc=3pt, before upper=\strut, top=-2pt, bottom=-4pt, left=-2pt, right=-2pt, boxrule=0pt}
\newtcbox{\hlsecondarytab}{on line, box align=base, colback=red!10,colframe=white,size=fbox,arc=3pt, before upper=\strut, top=-2pt, bottom=-4pt, left=-2pt, right=-2pt, boxrule=0pt}

\definecolor{darkred}{RGB}{200,0,0}
\definecolor{lightgreen}{RGB}{228,253,227}
\definecolor{lightred}{RGB}{252,231,234}
\definecolor{lightyellow}{RGB}{250,253,191}
\definecolor{lightblue}{RGB}{230,240,254}
\definecolor{white}{RGB}{255,255,255}

\newcommand{\authcomma}{\textmd{,}\hskip0.5em}

\usepackage[most]{tcolorbox}

\newcommand{\OurNameWithBold}{\textbf{Re}jection Sampling for Continued \textbf{Se}lf-instruction \textbf{T}uning}
\newcommand{\OurNameShort}{\textsc{ReSet}}

\title{Dancing in Chains: Reconciling Instruction Following \\and Faithfulness in Language Models}

\author{Zhengxuan Wu$^{1\diamond}$\authcomma Yuhao Zhang$^{2\diamond*}$\authcomma Peng Qi$^{3\diamond*}$\authcomma Yumo Xu$^{4*}$\authcomma\\ \textbf{Rujun Han}$^{5}$\authcomma
\textbf{Yian Zhang}$^{6\diamond}$\authcomma\textbf{Jifan Chen}$^{4}$\authcomma\textbf{Bonan Min}$^{4}$\authcomma\textbf{Zhiheng Huang}$^{7\diamond}$\\
$^1$Stanford University \quad$^2$Samaya AI \quad $^3$Orby AI \quad$^4$AWS AI Labs\\
\quad$^5$Google \quad$^6$NVIDIA \quad$^7$Denser.ai\\
\texttt{wuzhengx@stanford.edu} \quad\texttt{yuhao@samaya.ai}\quad \texttt{peng@orby.ai}\quad \texttt{yumomxu@amazon.com}\\
}

\begin{document}
\maketitle
\begin{abstract}

\renewcommand{\thefootnote}{\fnsymbol{footnote}}
\footnotetext[1]{Equal contribution. $^{\diamond}$Work done at AWS AI Labs.}
\renewcommand{\thefootnote}{\arabic{footnote}}

Modern language models (LMs) need to follow human instructions while being faithful; yet, they often fail to achieve both.
Here, we provide concrete evidence of a trade-off between instruction following (i.e., follow open-ended instructions) and faithfulness (i.e., ground responses in given context) when training LMs with these objectives. For instance, fine-tuning LLaMA-7B on instruction following datasets renders it less faithful. Conversely, instruction-tuned Vicuna-7B shows degraded performance at following instructions when further optimized on tasks that require contextual grounding. One common remedy is multi-task learning (MTL) with data mixing, yet it remains far from achieving a synergic outcome. We propose a simple yet effective method that relies on \OurNameWithBold{} (\textbf{\OurNameShort}), which significantly outperforms vanilla MTL. Surprisingly, we find that less is more, as training \textbf{\OurNameShort} with high-quality, yet substantially smaller data (three-fold less) yields superior results. Our findings offer a better understanding of objective discrepancies in alignment training of LMs.
\end{abstract}

\section{Introduction}
Aligning language models (LMs) with human preferences becomes increasingly important. One main objective is to train LMs to follow human instructions (e.g., answering open-ended questions)
while being faithful (e.g., grounding responses in the given context). However, LMs often suffer from failing to follow human instructions~\cite{mostlyknow,chen2023felm,ji2023survey} or making up facts that are not grounded in context~\cite{zhang2023siren,wang2023survey,hallucination_survey,ghosh2024closer}.

We trace this problem back to commonly used alignment training datasets, often collected from naturalistic conversations covering a wide range of domains~\cite{alpaca,koala_blogpost_2023,vicuna2023,christiano2017deep,ouyang2022training}. For instance, Alpaca~\cite{alpaca} and Dolly-15K~\cite{DatabricksBlog2023DollyV2} cover tasks from creative writing to context-dependent QA. Particularly, these tasks may have distinct objectives, and, when mixed, may induce potential conflicts of interest during alignment.

\begin{figure}[t]
 \centering
 \includegraphics[width=0.90\linewidth]{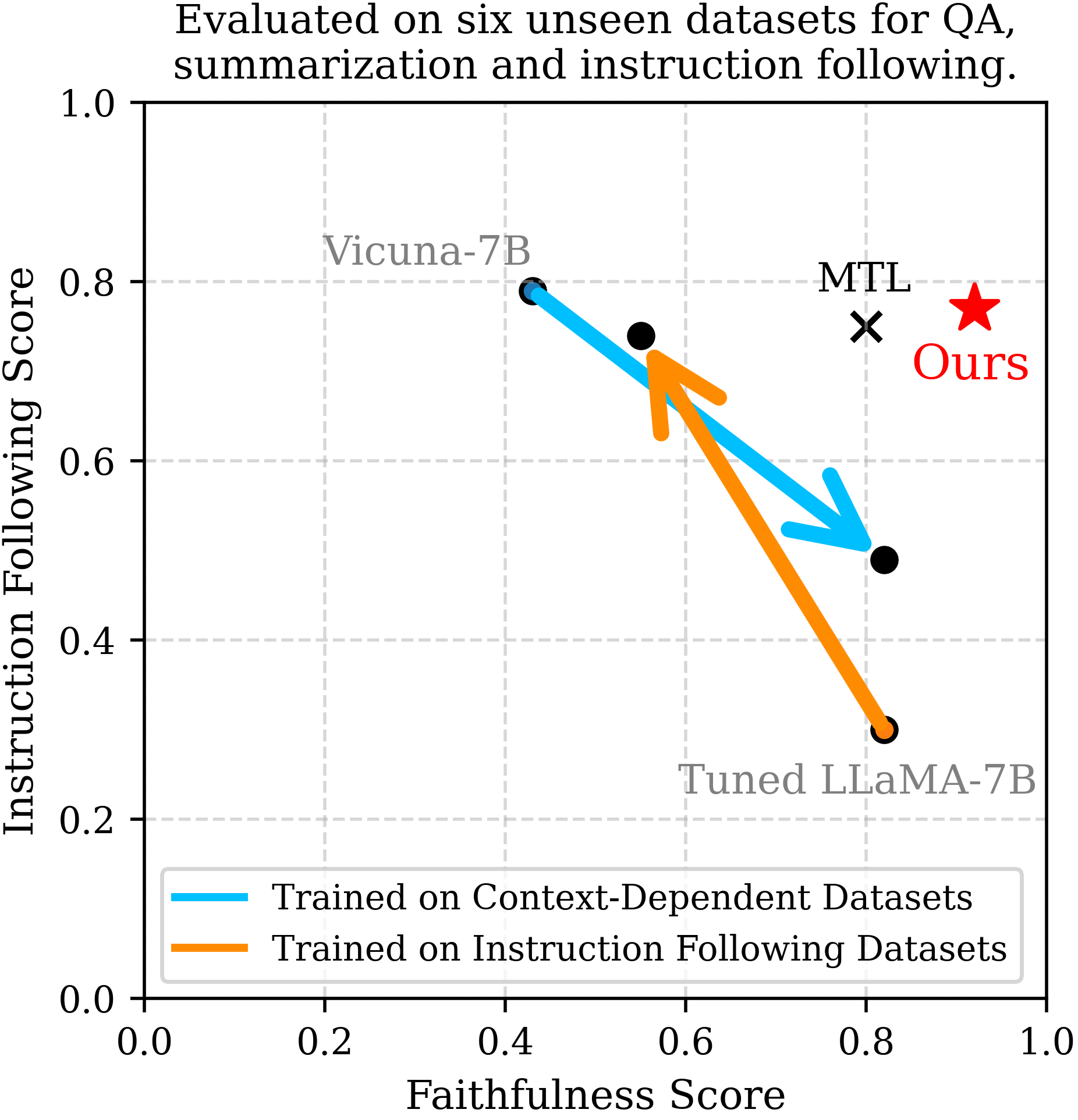}
 \caption{Faithfulness scores on context-dependent tasks (QA and summarization) decrease when we fine-tune grounded LLaMA-7B checkpoint with instruction following datasets (orange), and instruction following scores (assessed by GPT-4) decrease when we fine-tune Vicuna-7B with context-dependent tasks (blue). 
 Our method, \OurNameShort{} surpasses the vanilla MTL with data mixing, approaching the North Star (upper right corner).
 }
 \label{fig:main_figure_1}
\end{figure}

In this context, we examine the interaction between instruction following (i.e., how well does the LM follow open-ended instructions) and faithfulness (i.e., is the LM's response grounded in the context) during alignment training.
Specifically, we study how instruction following and faithfulness scores change when adapting LMs to two types of datasets: 1) \emph{instruction-tuning datasets} that are commonly used to train chat-models such as ChatGPT~\cite{openai2022intro} and Llama-2-Chat~\cite{touvron2023llama}; and 2) \emph{context-dependent datasets} that require grounding to a provided \emph{context}, and are commonly used to train Retrieval-Augmented Generation (RAG) models~\cite{lewis2020retrieval} such as Atlas~\cite{izacard2022atlas} and DSPy~\cite{khattab2023dspy}.
We observe a clear trade-off between the two scores, as shown in \Figref{fig:main_figure_1}: a fine-tuned LM with a competitive faithfulness score becomes much less faithful when separately fine-tuned on instruction following datasets. Conversely, an instruction-tuned LM becomes worse at following instructions when fine-tuned on context-dependent datasets. Our findings suggest fine-tuning an LM with either instruction following or context-dependent datasets exclusively may impair its original ability in the other aspect.

One natural mitigation strategy is to use multi-task learning (MTL) by mixing datasets, which we find as a strong yet sub-optimal baseline. 
To achieve a more synergic outcome, we propose a simple yet effective method \OurNameWithBold{} (\textbf{\OurNameShort}).
Inspired by recent works in self-instruct~\cite{zelikman2022star,selfinstruct}, \OurNameShort\ leverages the LM to sample generations for instruction following and task-specific datasets, which is different from vanilla MTL. Generations are then rated by external judges for instruction following and faithfulness scores, where top-rated generations are collected and used to further fine-tune the LM.
Our experiments show that when trained with \OurNameShort{} using only a single iteration and 8,000 additional fine-tuning examples, LMs see substantial gains in faithfulness scores (up to +18.8\%) compared to the MTL baseline, while maintaining their instruction-following scores. 
Furthermore, we find that less is more: training with \OurNameShort\ on higher quality yet three-fold less data yields up to 31.3\% improvements on faithfulness among datasets compared to MTL.
Our analyses shed lights on finding and mitigating objective discrepancies in alignment training where datasets encompass different or even conflicting goals.\footnote{We will release our code and evaluation data at\url{https://github.com/frankaging/dancing-in-chains}.}

\begin{table}[t]
    \centering
    \resizebox{1.0\columnwidth}{!}{
    \begin{tabular}{lrrrr}
    \toprule
    \textbf{Dataset} & \textbf{Train} & \textbf{Dev} & \textbf{Test} & \textbf{Avg. Length}\\
    \midrule
    {Instruction Datasets} & 257,307 & 2,500 & - & 305 \\
    \textit{Alpaca-15K} & - & - & 31,323 & 79 \\
    \textit{Vicuna-eval} & - & - & 80 & 367 \\
    \textit{Koala-eval} & - & - & 180 & 444 \\
    \midrule
    {NQ} & 69,639 & 8,757 & 13,368 & 6 \\
    {CNN/Daily Mail} & 287,113 & 13,368 & 11,490 & 76 \\
    {MS MARCO} & 153,725 & 2,500 & 12,466 & 23 \\
    \textit{BioASQ} & - & - & 1,956 & 9 \\
    \textit{SearchQA} & - & - & 31,760 & 4 \\
    \textit{WikiSum} & - & - & 2,000 & 140 \\
    \bottomrule
    \end{tabular}
    }
    \caption{Data statistics of two-stage fine-tuning experiments. 
    \textit{Italicized} datasets are held-out unseen evaluation datasets, while the rest are used for training. ``Instruction Datasets'' here consist of publicly avaliable datasets such as Dolly-15K~\cite{DatabricksBlog2023DollyV2}, ShareGPT, Self-Instruct~\cite{selfinstruct} and OASST-1~\cite{kopf2023openassistant}. Average length is the averaged response token length, which is calculated with training or testing sets for each dataset with the LLaMA-7B tokenizer.
    }
    \label{tab:data_stats}
\end{table}

\section{Background} 
To recap, our goal is to study the interplay of training on instruction following datasets and on datasets that require grounding to a specific context. 
To do this, we first construct two separate groups of datasets, each resembling a setting above.
We then adopt a two-stage fine-tuning paradigm that allows us to closely study the effect of each type of training.
We introduce the setup in this section.

\begin{table}[t]
    \centering
    \small
    \noindent\fbox{%
    \begin{minipage}{0.9\columnwidth} 
\tt 
Below is an instruction that describes a task, paired with an input that provides further context. Write a response that appropriately completes the request. \\

\#\#\# Instruction: \\
\textcolor{gray}{$[$Task-specific Instruction Abbreviated$]$} \\
\\
\#\#\# Input: \\
\textcolor{gray}{$[$Passage: Abbreviated$]$} \\
\textcolor{gray}{$[$... ...$]$} \\
\\
\#\#\#Response: \\
\textbf{\underline{Model Generated Answer Goes Here}}
    \end{minipage}
}
    \caption{The prompt template we used for training and evaluation for LLaMA-7B. The instruction field contains task-specific instruction, the input field contains contexts if applicable, and the response field is followed by model's generation. We use a different template for Vicuna-7B since it uses a different template during instruct-tuning phase. Actual input examples for each dataset and model are included in \Appref{app:exampele_instructions}.}
    \label{tab:instruction_template}
\end{table}

\begin{table*}
    \centering
    \resizebox{1.8\columnwidth}{!}{
    \begin{tabular}{p{3cm}p{13cm}}
    \toprule
    \textbf{Dataset Type} & \textbf{Instruction} \\
    \midrule
    \textbf{Extract QA} & Answer to the question by \textbf{extracting a specific text span} from the given passages. Do not include new information beyond the given passages. \\
    \midrule
    \textbf{Abstractive QA} & Answer the question with well-formed sentences. Paraphrase the context in the passages if necessary. \textbf{Do not include new information} beyond the given passages. \\
    \midrule
    \textbf{Summarization} & Summarize the text in a few sentences. Using original phrases or paraphrasing them if necessary. \textbf{Do not include new information} beyond the given passages. \\
    \bottomrule
    \end{tabular}
    }
    \caption{Task specific instructions. We bold texts that indicate that our prompts are designed to be objective-aligned with our instruction following training data (i.e., fine-tuning our model on instruction following datasets should with keeping it to be faithful as well). 
    }
    \label{tab:task_instruction}
\end{table*}

\subsection{Datasets}
We outline different datasets used for instruction following and context-dependent fine-tuning. \Tabref{tab:data_stats} shows data statistics. For each dataset we carefully design the instructions to well-align the input and output (see \Secref{sec:instr_template}).

\subsubsection{Instruction Following Datasets} 
We curated an instruction following training dataset by compiling unique examples from publicly available datasets such as Dolly-15K~\cite{DatabricksBlog2023DollyV2}, 
ShareGPT\footnote{\url{https://huggingface.co/datasets/LLMs/Alpaca-ShareGPT}}, Self-Instruct~\cite{selfinstruct}, and OASST-1~\cite{kopf2023openassistant}. For evaluation, we gather unique examples from Alpaca-15K~\cite{alpaca}, Vicuna-eval~\cite{vicuna2023}, and Koala-eval~\cite{koala_blogpost_2023}.

\paragraph{Data Pre-processing} We exclude examples that originally come with context (e.g., examples labeled as summarization type in Dolly-15K are filtered out) to prevent overlaps with our context-dependent datasets. We retain only unseen examples in our evaluation sets.
For OASST-1, we only include examples with an average human rating higher than 0.5 (i.e., the higher the rating is, the higher the quality is) and that are rated by at least two annotators. To refine ShareGPT, we include only examples with responses that are longer than 10 words, split by whitespace. For other instruction following training datasets, we exclude examples with empty instructions or responses.

\subsubsection{Context-Dependent Datasets}
We aim to evaluate the model's generation for faithfulness to the given input context. We select a range of context-dependent datasets from three task domains: (1) extractive QA including NQ~\cite{kwiatkowski-2019-natural}, BioASQ and SearchQA, taken from the RobustQA benchmark~\cite{han-2023-robustqa}; (2) abstractive QA with MS MARCO~\cite{bajaj2016ms} where the answers are well-formed sentences grounded in context; and (3) abstractive summarization including CNN DailyMail~\cite{herman2015teaching, nallapati-etal-2016-abstractive} and WikiSum~\cite{liu2018generating}. BioASQ, SearchQA and WikiSum are hold out for evaluation and the rest are for training. One crucial advantage of using these context-dependent datasets is that they provide us with a reliable way of measuring faithfulness, in terms of how well the response is grounded in the given context.

\paragraph{Data Pre-processing} For QA datasets, we include five retrieved passages maximally as the context,\footnote{Details about the retrieving process can be found in the original papers of RobustQA~\cite{han-2023-robustqa} and MS MARCO~\cite{bajaj2016ms}.} where the gold answer is at least mentioned in one of the passages. For MS MARCO, we only include examples where there exist at least one well-formed answers. In cases involving multiple retrieved passages, we concatenate all passages with line breaks inserted between them. 

\subsection{Evaluation Metrics}\label{sec:eval_metrics}
In the context of our datasets, we evaluate our models with three metrics: instruction following score, faithfulness score and task performance score. The standard methods for measuring instruction following and faithfulness of language models are subject to ongoing debate. In this work, we employ  widely adopted approaches for these measurements and compile a set of metrics for more stable evaluations.\footnote{Throughout the paper, we sample a subset of the full evaluation data which include 6,000 examples (1,000 examples from each context-dependent evaluation set), and sample 300 examples (100 examples from each instruction following evaluation set) due to limited compute resources.} To present our findings, we report macro-averaged results across all test datasets.

\paragraph{Instruction Following Score} We adopt the commonly used evaluation paradigm proposed by LLM-as-a-Judge~\cite{zheng2023judging}, and zero-shot prompt GPT-4 to provide a rating followed by a short explanation (i.e., named as LLM-as-a-Judge (R) in the paper).\footnote{We use \texttt{gpt-4-0613}.} For the GPT-4 evaluator, we set the temperature to 0 for stability with a maximum generation length of 512. We check instruction following scores only for instruction following evaluation datasets. See \Appref{app:eval_human} for our actual evaluation prompt.

\paragraph{Faithfulness Score} For extractive QA datasets, we utilize the span coverage score as our metric (i.e., whether the predicted answer is a span within the context). We apply standard normalization to both the predicted answer and the context (see \Appref{app:eval_faith} for details). A score of 1.0 is assigned if the span is covered, and 0.0 otherwise. Additionally, we include unigram and bigram coverage for selected datasets to further refine our faithfulness evaluation in the Appendix (see  \Figref{fig:paper_quandary_stage_1} and \Figref{fig:paper_qa_summ_stage_2}). For abstractive QA and summarization datasets, we employ SummaC-ZeroShot (SummaC-ZS;~\citet{Laban2022SummaCRN}) to assess whether the provided context (with the question concatenated as a prefix for QA datasets) entails the model-generated answer. Specifically, we segment both the context and the answer into sentences and then use a pretrained NLI model to compute an entailment score between pairs. These NLI scores are aggregated across pairs using max-pooling.

\paragraph{Task Performance} Task performance complements the aforementioned metrics, as task performance is generally expected to decline when models are fine-tuned across domains. We utilize Exact Match (EM)
for extractive QA datasets and the ROUGE-L score~\cite{lin-2004-rouge}
for abstractive QA and summarization datasets.

\subsection{Instruction Template}\label{sec:instr_template}

Our datasets follow the same instruction-tuning template format as used in the Alpaca setting~\cite{alpaca}. This template includes a header that outlines general guidelines, followed by task-specific instructions, as illustrated in \Tabref{tab:instruction_template}. For QA tasks, the question is presented after the retrieved passages.

\paragraph{Objective-Aligned Instructions} We design our task-specific instructions to minimize \emph{objective-conflict} among datasets. For example, in prompting our model for a context-dependent extractive QA task, we explicitly instruct the model to ``extract a specific text span from the given passages''. This template helps models to reduce hallucination when fine-tuned with instruction following datasets: as models getting better at understanding human instructions, they also get better at understanding how to \emph{extracting a span} which ensures the answer to be faithful. The instructions for all datasets are depicted in \Tabref{tab:task_instruction}. Our task-specific templates ensure there is no objective conflict when we fine-tune our models with mixed of datasets. We use the same template for training and evaluation for each task.

\begin{figure}[t]
 \centering
 \includegraphics[width=0.90\linewidth]{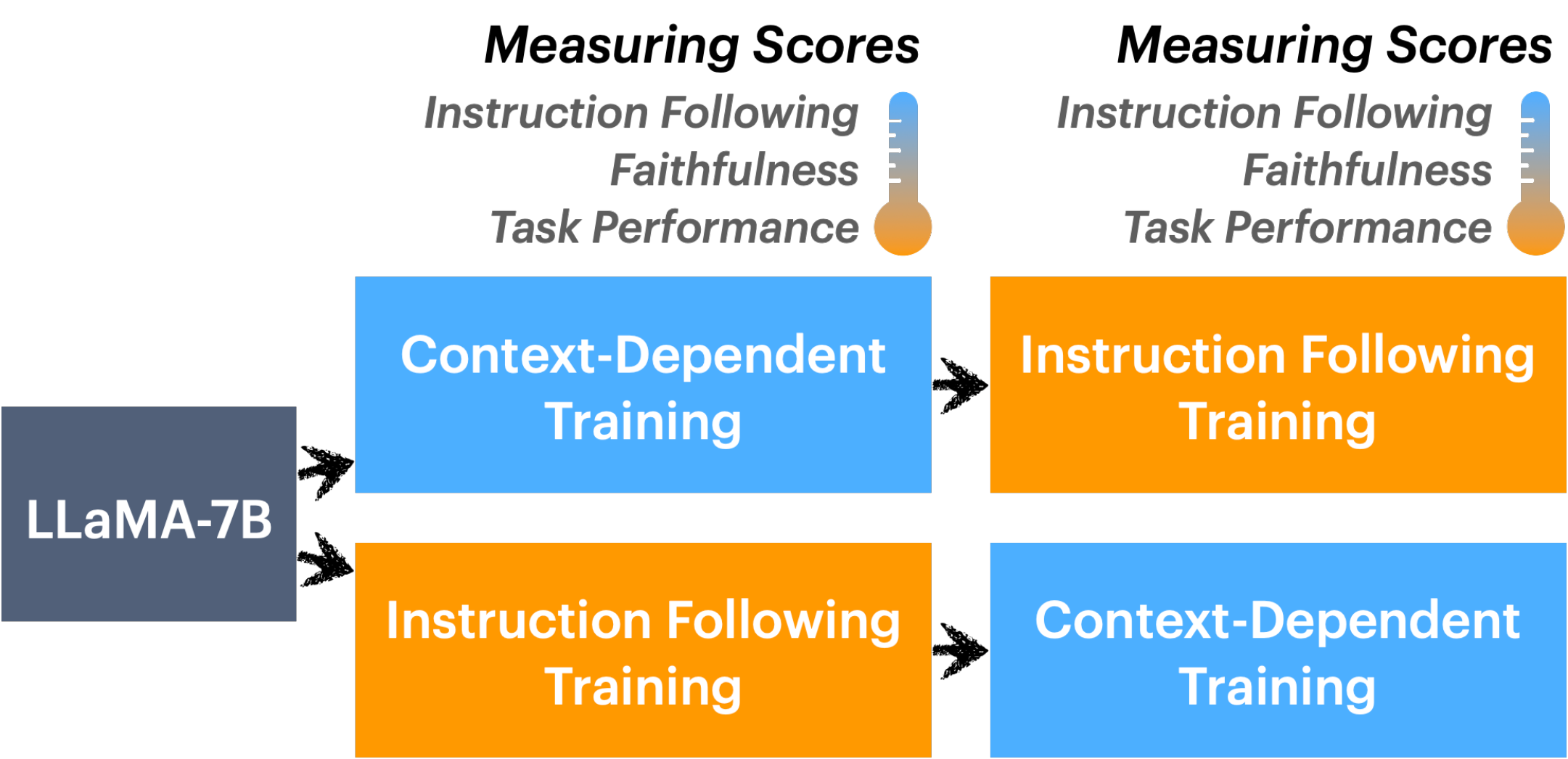}
 \caption{Two-stage fine-tuning with LLaMA-7B.
 }
 \label{fig:inoculation_diagram}
\end{figure}

\begin{figure*}[t]
 \centering
 \includegraphics[width=1.0\linewidth]{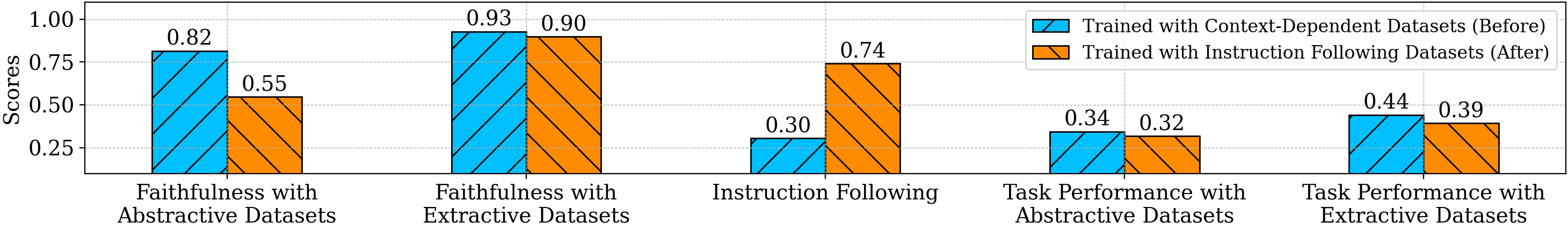}
 \caption{Macro-averaged faithfulness, instruction following, and task performance scores on corresponding evaluation datasets before and after fine-tuning with instruction following datasets.}
 \label{fig:RQ_1_result_1}
\end{figure*}

\begin{figure*}[t]
 \centering
 \includegraphics[width=1.0\linewidth]{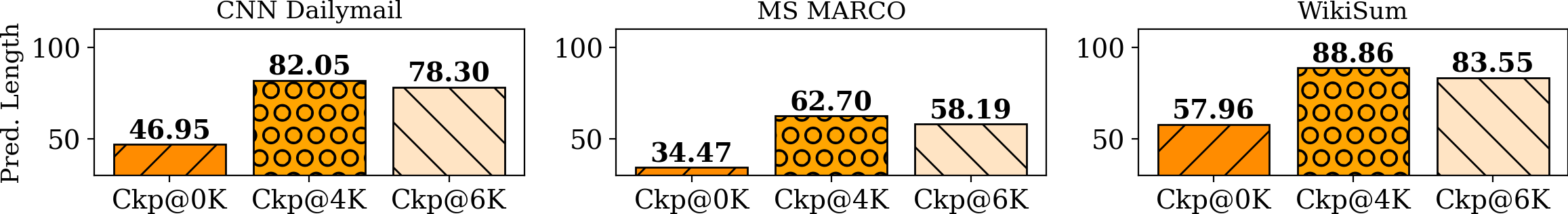}
 \caption{Average generation token length throughout the instruction following training stage. The first checkpoint is the best checkpoint from the context-dependent training stage. The middle checkpoint is with the lowest evaluation loss during the second stage.}
 \label{fig:paper_seq_length_abstract_stage_2}
\end{figure*}

\begin{figure}[t]
 \centering
 \includegraphics[width=0.9\linewidth]{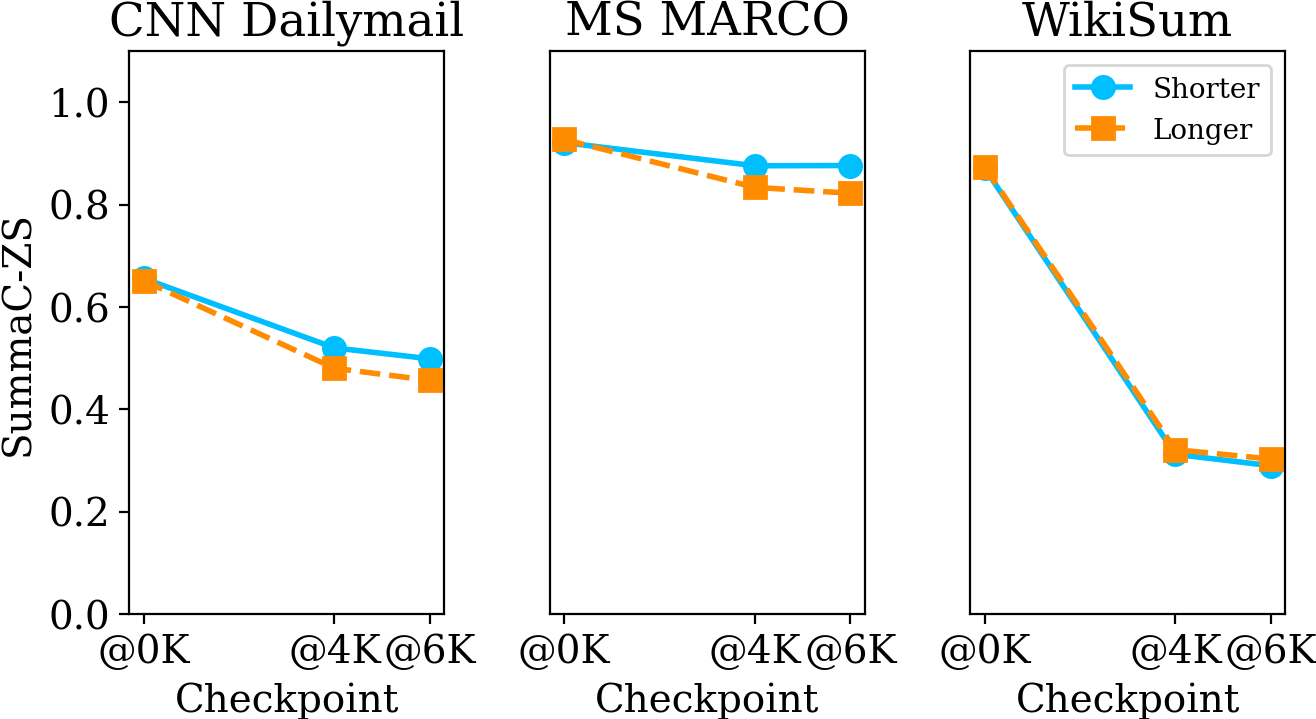}
 \caption{Faithfulness scores for abstraction QA and summarization datasets categorized by whether the generation length is strictly shorter or much longer ($>$ 100 tokens) than the golden answer.}
 \label{fig:paper_abstract_length_split_stage_2}
\end{figure}

\subsection{Two-stage Fine-tuning Paradigm}
To understand the trade-off between instruction following and faithfulness when training LMs with both objectives, we formulate a two-stage fine-tuning paradigm to answer our research questions (as illustrated in \Figref{fig:inoculation_diagram}).\footnote{Our approach is akin to the data inoculation paradigm proposed by \citet{liu-schwartz-smith:2019:NAACL}, albeit with significantly larger models and datasets.} For our first pipeline, we initially fine-tune our LM with context-dependent datasets that require grounding. We then take the best checkpoint from the initial stage to further fine-tune it on instruction following datasets (CD$\to$IF). Conversely, in our second pipeline, we fine-tune instruct-tuned LM (e.g., Vicuna-7B) with context-dependent datasets (IF$\to$CD). For both pipelines, we measure instruction following and faithfulness scores before and after training, to gauge the impact of the second-stage training on both capabilities. We follow this paradigm to find evidence of the trade-off in \Secref{sec:finetune_instru} and \Secref{sec:finetune_ctx}.

\paragraph{Models} We use two models in our two-stage fine-tuning paradigm. We use a base LM LLaMA-7B~\cite{touvron2023llama}, one of the most widely used open-source LM, for the CD$\to$IF pipeline. For our IF$\to$CD pipeline, we use Vicuna-7B off-the-shelf as our instruct-tuned LLaMA-7B without retraining one from scratch. Vicuna-7B is one of the most competitive open-source chat-model, and is a fine-tuned LLaMA-7B on conversational data from ShareGPT~\citep{zheng2023judging}.
Our hypothesis and paradigm are transferable to other base LMs at different scales, although we pick these two models as they are among the first open-sourced LMs during the time frame of this project. Other experimental setup details are included in \Appref{app:setup}.

\begin{figure*}[t]
 \centering
 \includegraphics[width=1.0\linewidth]{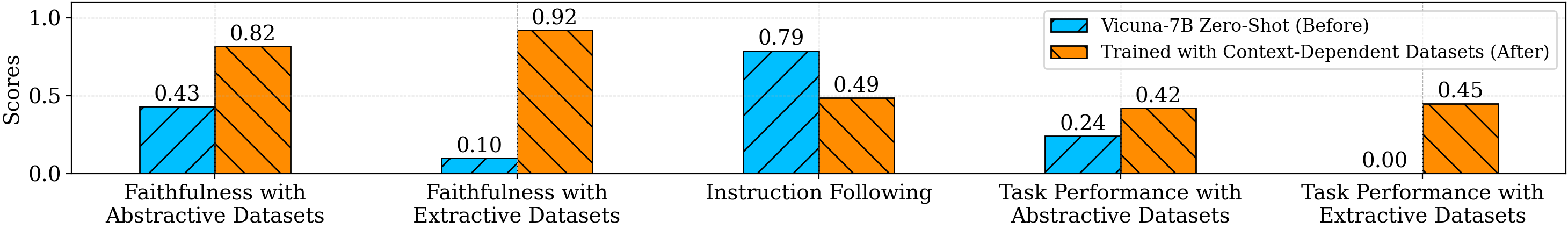}
 \caption{Macro-averaged faithfulness, instruction following, and task performance scores on corresponding evaluation datasets before and after fine-tuning with context-dependent datasets.}
 \label{fig:RQ_2_result_1}
\end{figure*}

\begin{figure}[t]
 \centering
 \includegraphics[width=1.0\linewidth]{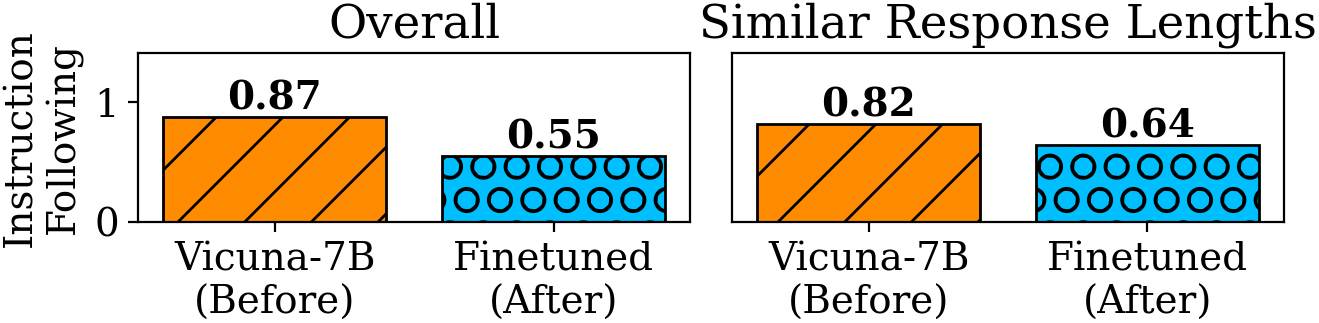}
 \caption{Instruction following scores for 1,000 randomly selected examples from Alpaca-15K (left), and for filtered examples with similar generation lengths (maximally 10 tokens longer) before and after fine-tuning with context-dependent datasets (right).}
 \label{fig:paper_alignment_length_split_stage_2}
\end{figure}

\section{Does Fine-tuning with Instruction Following Data Hurt Faithfulness?}\label{sec:finetune_instru} 

To answer this question, we follow the first pipeline outlined in Figure \ref{fig:inoculation_diagram}, where we take our LLaMA-7B that is fine-tuned on context-dependent datasets and further fine-tune it with instruction following datasets. Our results are shown in Figure \ref{fig:RQ_1_result_1}. First of all, instruction following scores increase drastically as expected, from 0.30 to 0.74. Meanwhile, faithfulness scores on abstractive datasets drop 33.0\% from 0.82 to 0.55, while the task performance is largely maintained (from 0.34 to 0.32). For extractive datasets, both faithfulness scores and task performance see a relatively small drop. To provide a fine-grained view of these datasets, we provide additional analysis of per-task and per-metric changes of faithfulness score as well as task performance in \Appref{app:additional_alignment_t} and \Appref{app:quandary_of_mix}.

One potential confounding factor for faithfulness scores dropping on abstractive tasks is the length of the generated response.
The model could simply have generated longer responses (\Figref{fig:paper_seq_length_abstract_stage_2}) as a result of training on intruction tuning data with longer responses (Table \ref{tab:data_stats}).
To rule this out, we re-evaluate our models separately with two contrasting groups: evaluated with only those examples with shorter generation length (less than or equal to), and those examples with much longer generation length (100 more tokens) compared with the golden answer. As shown in Figure \ref{fig:paper_abstract_length_split_stage_2}, both short and long generations see very substantial drops in faithfulness, while longer generations indeed see larger drops. This nevertheless supports our conclusion that instruction following training hurts faithfulness.

\begin{figure*}[t]
 \centering
 \includegraphics[width=1.0\linewidth]{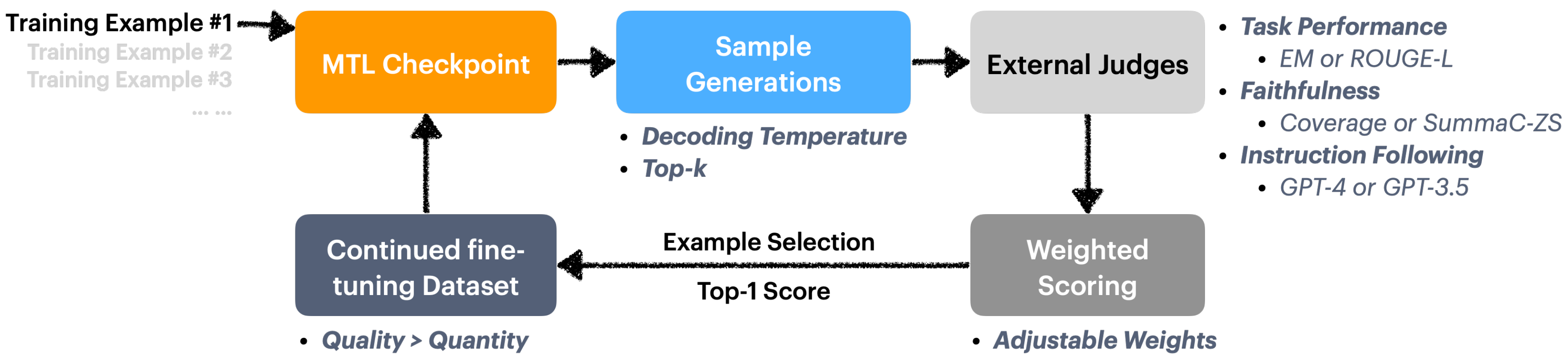}
 \caption{The illustration of our proposed method \OurNameShort{}. It samples generations from the initial vanilla multi-task learning (MTL) checkpoint with seen examples from instruction following and context-dependent datasets. For each example, it generates a set of possible responses with different decoding strategies. Generations are rated by external judges with a weighted scores of task performance, faithfulness and instruction following scores. Then, the top rated generations will be collected to further fine-tune the initial model.}
 \label{fig:our_method_diagram}
\end{figure*}

\section{Does Context-Dependent Fine-tuning Hurt Instruction Following?} \label{sec:finetune_ctx} 
To answer this question, we fine-tune Vicuna-7B (LLaMA-7B fine-tuned on ShareGPT) on context-dependent datasets. Figure \ref{fig:RQ_2_result_1} shows our results. As expected, our model becomes more faithful, with improved task performance. This improvement is partly because Vicuna-7B does not follow our instruction template out-of-the-box. On the other hand, instruction following scores drop by 37.9\%, suggesting our model becomes much less human-aligned. Similar to previous findings, one natural concern is that context-dependent training makes our model generate much shorter responses, which leads to lower instruction following scores. To rule out this concern, we randomly sample 200 examples from the Alpaca-15K dataset and only keep those with a minimal sequence length change (a maximum of 10 new tokens) compared to the corresponding Vicuna-7B zero-shot generation. As shown in Figure \ref{fig:paper_alignment_length_split_stage_2}, instruction following scores consistently drop considerably for the filtered setting, regardless of length (36.8\% vs.\ 22.0\%). Overall, context-dependent fine-tuning negatively affects the instruction following score, as we expected.

\section{\textbf{\OurNameShort}: Reconciling Instruction Following and Faithfulness}
As shown in the previous two sections, there are clear objective discrepancies between diverse instruction following datasets and context-grounded ones. In this section, we explore two different mitigation strategies, starting from the MTL baseline to our novel method based on \OurNameWithBold{} (\textbf{\OurNameShort}), which significantly outperforms MTL.

\paragraph{Our MTL Baseline} Our MTL simply mixes up the instruction following as well as context-dependent datasets. We up-sample smaller datasets to have an equal number of examples per dataset.

\subsection{Our Method}
\OurNameShort{} starts with a model that is fine-tuned on a mixture of instruction following and context-dependent datasets and reconciles the two objectives with the following steps (\figref{fig:our_method_diagram}): 

\paragraph{Sample Generations} For a random subset of the training datasets, we sample generations from the checkpoint with different decoding settings. Specifically, we focus on two hyperparameters by randomly changing one of them at a time by enumerating all possible values. For the decoding temperature, it takes on a value from \{0.1, 0.2, 0.3, 0.4, 0.5, 0.6, 0.7\}. For top-k, we set it to take on a value from \{5, 10, 20, 50, 70, 90, 100\}. When we vary the temperature, we fix top-k to be 0. Similarly, when we vary the top-k, we fix the temperature at 0. For each example, if we run the sampling procedure once, it will sample 7 examples in total.

\paragraph{External Judges} We then use a set of external judges to rate our collected generations as in \Secref{sec:eval_metrics}. These judges are offline evaluators. We evaluate generation based on instruction following score, faithfulness score and task performance. One potential limitation is that our model may overfit to existing judges on seen datasets. We thus include unseen datasets to test generalizability.

\paragraph{Top-1 Weighted Score} For each generation, we collect a set of scores from our external judges. Then, we take a weighted sum of these scores before ranking the generations and picking the top rated one. The score is weighted score of task performance $s_{\text{task}}$, instruction following $s_{\text{instr}}$ and faithfulness scores $s_{\text{faith}}$,
\[
\mathrm{score}=s_{\text{task}}+2.0*(\mathbb{I}_{\text{instr}}*s_{\text{instr}} + \mathbb{I}_{\text{faith}}*s_{\text{faith}})
\]
where we use $\mathbb{I}_{\text{instr}}$ and $\mathbb{I}_{\text{faith}}$ to indicate whether the example is from our instruction following datasets or context-dependent datasets.
We pick the top rated sample per example and combine them into a continued fine-tuning dataset as described next.

\paragraph{Continued Fine-tuning} For \OurNameShort{}, we randomly sample training data from each of our four dataset in \Tabref{tab:data_stats} and collect 2,000 additional fine-tuning examples per dataset. In total, our new fine-tuning dataset has 8,000 examples. With this small collected dataset, we further fine-tune our starting checkpoint model for a single epoch with a smaller learning rate to avoid overfitting. This continued fine-tuning step is very lightweight as the training data is usually less than 1\% of the MTL training step. Overall, \OurNameShort\ resembles a reject sampling based preference learning paradigm, which has been proven effective~\cite{touvron2023llama} while drastically saving training costs as well as increasing stability. Other experimental setup details are included in \Appref{app:setup}.

\paragraph{Supercharged \OurNameShort{} (\OurNameShort{}-S)}
In addition, we evaluate \OurNameShort{} with a different setting to test the impact of the quality of our continued fine-tuning dataset. Specifically, we \emph{supercharge} the quality of our additional fine-tuning dataset by sampling 1$\times$ more generations. We swap our instruction following judge from GPT-4 to the weaker ChatGPT\footnote{We use \texttt{gpt-4-0613}.}. One potential benefits is that ChatGPT is a weaker judge with a lower recall for good generations. As a result, examples rated high by ChatGPT may have higher quality.
While sampling for generation, we decrease the curated fine-tuning dataset by 3-fold (2,000 examples in total). We fine-tune our checkpoint model with the same setup.

\begin{table*}[t]
\resizebox{1.00\linewidth}{!}{%
\begin{tabular}{lccccccccccc}
\toprule
 & \multicolumn{9}{c}{\textbf{Faithfulness}} & & \multicolumn{1}{c}{\textbf{Instruction Following}} \\ 
 
 & \multicolumn{4}{c}{\textbf{Abstractive Datasets}} & & \multicolumn{4}{c}{\textbf{Extractive Datasets}} & & \multicolumn{1}{c}{\textbf{Datasets}} \\ \midrule
 & CNN DailyMail & MS MARCO & \textit{WikiSum} & Overall & & NQ & \textit{SearchQA} & \textit{BioASQ} & Overall & & \textit{Alignment Eval} \\ 
 \cmidrule(lr){1-5} \cmidrule(lr){6-10} \cmidrule(lr){11-12}
 \textbf{Vicuna-7B} & 0.36 & 0.59 & 0.34 & 0.43 & & 0.06 & 0.21 & 0.02 & 0.10 & & 0.79 \\ 
 { }\textbf{w/o MTL} & 0.67 & \textbf{0.93} & 0.85 & 0.82  & & 0.89 & 0.94 & 0.93 & 0.92 & & 0.49 \\ 
 { }\textbf{w/ MTL} & 0.67 & 0.92 & 0.80 & 0.80 & & 0.90 & 0.94 & 0.93 & 0.92 & & 0.75 \\ 
 \midrule
 { }\textbf{w/ \OurNameShort{}} & 0.80 & 0.90 & 0.88 & 0.86 & & \textbf{0.96} & \textbf{0.99} & 0.92 & 0.96 & & 0.73 \\ 
 { }\textbf{w/ MTL+\OurNameShort{}} & 0.77 & \textbf{0.93} & \textbf{0.95} & 0.85 & & \textbf{0.96} & 0.97 & \textbf{0.98} & \textbf{0.97} & & 0.76 \\ 
 { }\textbf{w/ MTL+\OurNameShort{}-S} & \textbf{0.88} & \textbf{0.93} & \textbf{0.95} & \textbf{0.92} & & 0.94 & 0.97 & 0.97 & 0.96 & & \textbf{0.77} \\ 
\bottomrule
\end{tabular}}
\caption{Faithfulness and alignment scores on testing datasets, and unseen datasets are \textit{italicized}.
Scores are averaged across three distinct runs. Higher scores are better. Overall scores are macro-average across datasets.}
\label{tab:our_method_result}
\end{table*}

\subsection{Results}
\Tabref{tab:our_method_result} shows results on models fine-tuned with other baselines: 
(1) \textbf{Vicuna-7B}: Evaluated with Vicuna-7B in a zero-shot manner. 
(2) \textbf{w/o MTL}: We fine-tune Vicuna-7B with context-dependent datasets without any MTL objective. The purpose is to establish a potential upper bound of faithfulness score without any further training for instruction following.
(3) \textbf{w/ MTL}: We fine-tune Vicuna-7B with baseline MTL by mixing instruction following and context-dependent datasets together. 
(4) \textbf{w/ \OurNameShort{}}: We fine-tune Vicuna-7B directly with our collected continued fine-tuning dataset (8,000 examples) from our \OurNameShort{} pipeline without MTL training first. Note that the curated dataset comprises of sampled generations from our MTL checkpoint. The purpose is to evaluate whether MTL is necessary for model improvements.
(5) \textbf{w/ MTL+\OurNameShort{}}: We follow \Figref{fig:our_method_diagram} to fine-tune our MTL checkpoint with curated dataset.
(6) \textbf{w/ MTL+\OurNameShort{}-S}: We follow \Figref{fig:our_method_diagram} to fine-tune our MTL checkpoint with the high-quality version of our curated dataset (2,000 examples).

We evaluate our models on both seen and unseen testing datasets. As shown in \Tabref{tab:our_method_result}, fine-tuning Vicuna-7B with a mixture of datasets close the gap on instruction following score substantially (from 0.49 to 0.75) while leaving headroom across the board. Next, models fine-tuned with \OurNameShort{} significantly outperform the MTL baseline on both seen and unseen testing datasets. Our results also suggest directly fine-tuning Vicuna-7B performs worse compared to fine-tuning a model checkpoint after MTL training. Last but not least, our results with \OurNameShort{}-S provide strong evidence that data quality is more important than data quantity and that using multiple iterations of sampling helps, as \OurNameShort{}-S achieves similar or better performance with 3-fold less training data. Due to space limit, we present qualitative model generations in \Appref{app:example_generation}.

\section{Related Work}

\paragraph{Instruction Following of LMs} 
There are a variety of instruction following training datasets~\cite{alpaca,koala_blogpost_2023,vicuna2023,selfinstruct} covering tasks from wildly different domains such as poetry creation to context-dependent summarization. In this paper, we focus on a particular trade-off between instruction following and model faithfulness, which are fundamental objectives for modern LMs. In addition to training datasets, LM generations are often assessed by human experts~\cite{ouyang2022training} or model scorers~\cite{zheng2023judging,pmlr-v162-ethayarajh22a}. We use GPT-4 to access instruction following scores as in \citet{vicuna2023}.

\paragraph{Faithfulness and Groundedness of LMs} Being faithful or grounded is crucial for tasks like context-dependent QA or summarizations, reducing hallucinations for LMs~\cite{zhang2023siren,wang2023survey,hallucination_survey,mostlyknow,chen2023felm,ji2023survey}
Unlike task performance, faithfulness measures whether generated answers are based on the given context~\cite{Rashkin_2021, dziri2022origin, paranjape2022hindsight}. For context-dependent QA and summarization benchmarks, common metrics include subsequence-based lexical matching~\cite{papineni-2002-bleu, lin-2004-rouge, banerjee-2005-meteor}, natural language inference \citep[NLI; ][]{Laban2022SummaCRN, fabbri-2022-qafacteval}, and more recently, LLMs as factuality scorers~\cite{chiang2023large,liu-2023-g,kamalloo2023evaluating}. We use lexical matching and NLI to measure faithfulness by checking whether the answer is grounded.

\paragraph{Instruction Following Training with LMs} Base LMs are often tuned to follow human instructions~\cite{wei2021finetuned,Mishra_2022,Wang_2022,chung2022scaling}. Various post-hoc fine-tuning techniques have been proposed to align base LMs with human preferences, such as supervised fine-tuning on instruction-following datasets~\cite{alpaca,vicuna2023} or variances of RLHF and RLAIF~\cite{schulman2017proximal,ouyang2022training,lee2023rlaif,rafailov2023direct,touvron2023llama}. These techniques can require training policy and reward models which is not cost-efficient. Inspired by alignment training with reject-sampling~\cite{touvron2023llama}, we propose a simple yet effective method that uses reject-sampling by self-instruct. On the other hand, there exist various advance MTL techniques~\cite{liu-2019-multi,crawshaw2020multi} prior to the development of LMs. However, \OurNameShort\ is the only variant that leverages LM's self-instruct generations to further fine-tune the LM. More importantly, \OurNameShort\ is complimentary to different instruct-tuning techniques by leveraging the LM and external evaluators to generate high-quality continued fine-tuning datasets. 
Compared with recent works showing how instruct-tuning may cause hallucination~\cite{ghosh2024closer}, we provide a more rigorous analysis with our two-stage fine-tuning paradigm and focus on the interaction between instruction following and faithfulness.

\section{Conclusion}
In this paper, we demonstrate a clear trade-off between instruction following and faithfulness when fine-tuning LMs on datasets with differing objectives.
To alleviate this trade-off, we proposed \textbf{\OurNameShort}, a simple yet effective iterative method that significantly outperforms the MTL baseline in both instruction following and faithfulness scores. The lightweight and iterative nature of \OurNameShort{} makes it extensible for future refinement at minimum cost and integrable with recent instruct-tuning techniques. We contribute to the broader goal of creating more reliable, accurate, and user-aligned language technologies.

\section*{Limitations} 
The limitations of our work are as follows:
\begin{itemize}
  \item Our study primarily focuses on the LLaMA-7B and Vicuna-7B models, that are among the best open-source models at the time of this work. While we posit that our findings and the proposed \OurNameShort{} method could generalize across other language models, our findings could remain speculative without evaluating on more current model types (e.g., LLaMA-2, Mistral or Mixtral at various scales).
  \item The datasets chosen for fine-tuning and evaluation, though comprehensive, are not exhaustive. There are other interesting datasets that are not covered in this study. For instance, long-form QA where the answers are much longer than 1-2 phrases or sentences. Our instruction following datasets can also be further categorized into creativity-driven, world-knowledge driven and others to help us to disentangle discrepancies in objectives better across datasets.
  \item Our evaluation relies heavily on automated metrics and external judges like external LMs for assessing instruction following and faithfulness. While these methods are standard, they cannot fully encapsulate the nuanced understanding and preferences of human evaluators. For future research, evaluating responses with human annotators would provide additional validations.
  \item Although the purpose of our study is to study the objective discrepancies in the datasets and come up with mitigation strategies without another novel training paradigm, it would strengthen our results if we can compare our method with more recent alignment training methods.
  \item Although we evaluate \OurNameShort{} on unseen datasets, our method still has the potential to overfit to certain evaluators. Future work may use a different set of evaluators for a more robust evaluation. Human evaluation is challenging for our unseen testing datasets, such as WikiSum, because the input and response are extremely long, which could also make human ratings unstable.
\end{itemize}

\section*{Ethics Statement}
In this paper, we delve into the nuances of instruct-tuning LMs, a process that involves balancing a trade-off between instruction following and faithfulness. We feel that the immediate ethical and societal consequences of our research may be limited, and none which we feel must be specifically highlighted here.

\bibliography{custom}

\appendix

\section{Appendix}
\label{sec:appendix}

\subsection{Evaluation Metric: Instruction Following Score}\label{app:eval_human}
\Tabref{tab:gpt4_eval_prompt} shows our actual prompt template for evaluation. We follow the prompt template provided in the LLM-as-a-Judge~\cite{zheng2023judging} paper. We normalize the return rating to [0, 1].

\begin{table*}[ht]
    \centering
    \small
    \noindent\fbox{%
    \begin{minipage}{2.0\columnwidth} 
\tt 
$[$System$]$\\
Please act as an impartial judge and evaluate
the quality of the response provided by an AI assistant to the user question displayed
below. Your evaluation should consider factors such as the helpfulness, relevance,
accuracy, depth, creativity, and level of detail of the response. Begin your evaluation
by providing a short explanation (strictly 1-2 short sentences). Be as objective as possible. After providing your
explanation, please rate the response on a scale of 1 to 10 by strictly following
this format: ``$[[$rating$]]$'', for example: ``Rating: $[[$5$]]$''. \\ \\
$[$Question$]$ \\ 
\textcolor{gray}{$[$The Question Abbreviated$]$} \\

$[$The Start of Assistant’s Answer$]$ \\
Answer: \textbf{\underline{Model Generated Answer Goes Here}}\\
$[$The End of Assistant’s Answer$]$

    \end{minipage}
}
    \caption{The template of the prompt we used for evaluating instruction following scores using GPT-4. This template is adopted from LLM-as-a-Judge~\cite{zheng2023judging} paper.}
    \label{tab:gpt4_eval_prompt}
\end{table*}

\begin{table}[t]
    \centering
    \small
    \noindent\fbox{%
    \begin{minipage}{1.0\columnwidth} 
\tt 
A chat between a curious user and an artificial intelligence assistant. The assistant gives helpful, detailed, and polite answers to the user's questions. \\

\#\#\# USER: \textcolor{gray}{$[$Task-specific Instruction Abbreviated$]$} \\
\textcolor{gray}{$[$Passage: Abbreviated$]$} \\
\textcolor{gray}{$[$... ...$]$} \\
\#\#\#ASSISTANT: \textbf{\underline{Model Generated Answer Goes Here}}
    \end{minipage}
}
    \caption{The prompt template we used for training and evaluation for Vicuna-7B. The instruction field contains task-specific instruction, the input field contains contexts if applicable, and the response field is followed by the model generation.}
    \label{tab:vicuna_instruction_template}
\end{table}

\subsection{Evaluation Metric: Faithfulness Score}\label{app:eval_faith}

\paragraph{Text Normalization} We use regular expressions to replace spaces around hyphens, slashes, and before ``'s''. We then remove all the articles (e.g., ``a'' and ``the'') and punctuations. We lowercase all letters for simplicity.

\paragraph{Common Failure Mode} The authors also human label model's generation to check whether model's generations are paraphrased version of the golden answers. The most common failure mode is the model extracting a non-existent span or a wrong span. This supports our findings of models being unfaithful.

\subsection{Experimental Setup}\label{app:setup}
We train our model for a maximum of a single epoch\footnote{Experiments with up to three epochs showed minimal changes in results.} across all training jobs. We up-sample the smaller datasets to match the number of examples in the larger ones when combing datasets for training. The learning rate is set at $1\times 10^{-5}$ with a batch size of 16 for our faithfulness-driven training. For training jobs involving instruction following datasets, we increase the batch size to 32. Training is conducted using \texttt{bfloat16} precision with a maximum sequence length of 2048. The weight decay is set to $0.05$, with a \texttt{cosine} learning rate scheduler and a warm-up ratio of $0.03$. We save checkpoints every 100 training steps and evaluate them based on perplexity scores on the evaluation set. The best-performing checkpoint is then selected for the next stage of training. Our models are trained using the stage-3 \texttt{deepspeed} library. We train each model with three random seeds and average the results for consistency. For each training job, our models are trained on 8$\times$A100 GPUs within a single-node setup, with the total training time not exceeding 24 hours. For model generation, we employ greedy decoding with a maximum generation length of 480, which aligns with the maximum response length across the training datasets.

For the contined fine-tuning step in \OurNameShort{}, we use a smaller learning rate of $8\times10^{-6}$ and keep other settings the same.

\subsection{Additional Analysis on Instruction Following Fine-tuning}\label{app:additional_alignment_t}

\begin{figure*}[t]
 \centering
 \includegraphics[width=1.0\linewidth]{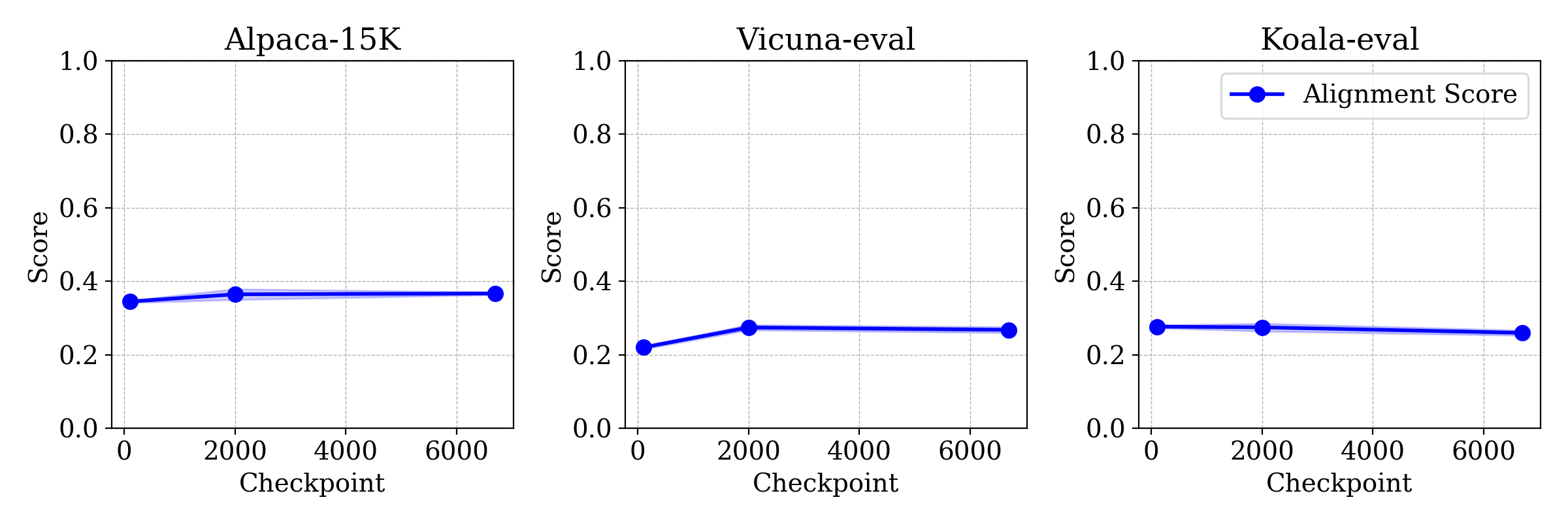}
 \caption{Individual instruction following score on alignment datasets evaluated with three distinct model checkpoints of LLaMA-7B fine-tuned on context-dependent datasets. The middle checkpoint is the one with lowest in-training evaluation loss.}
 \label{fig:paper_alignment_stage_1}
\end{figure*}

\begin{figure*}[t]
 \centering
 \includegraphics[width=1.0\linewidth]{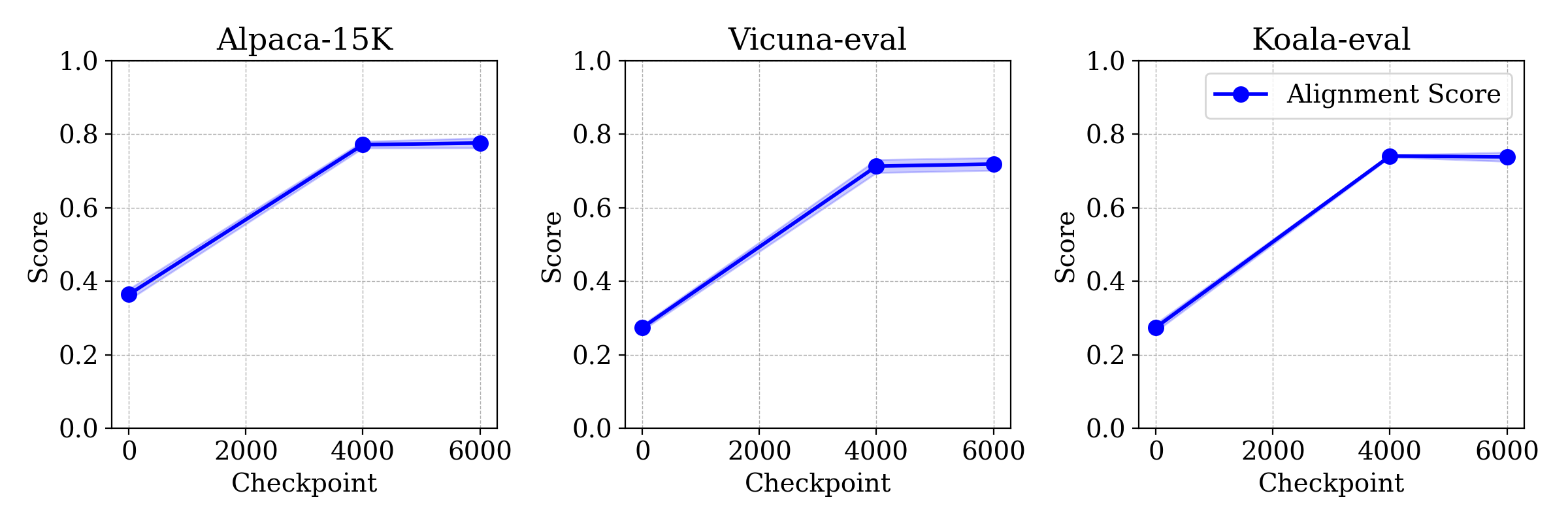}
 \caption{Individual instruction following score on alignment datasets evaluated with three distinct model checkpoints through the instruction following fine-tuning. The middle checkpoint is the one with lowest in-training evaluation loss.}
 \label{fig:paper_alignment_stage_2}
\end{figure*}

\Figref{fig:paper_alignment_stage_1} shows how instruction following scores vary during fine-tuning LLaMA-7B with context-dependent tasks. \Figref{fig:paper_alignment_stage_2} shows how instruction following scores vary during the second-stage fine-tuning with instruction following datasets.

\subsection{Quandary of Mixed Training on Abstractive and Extractive QA and Summarization Datasets}\label{app:quandary_of_mix}

\begin{figure*}[t]
 \centering
 \includegraphics[width=1.0\linewidth]{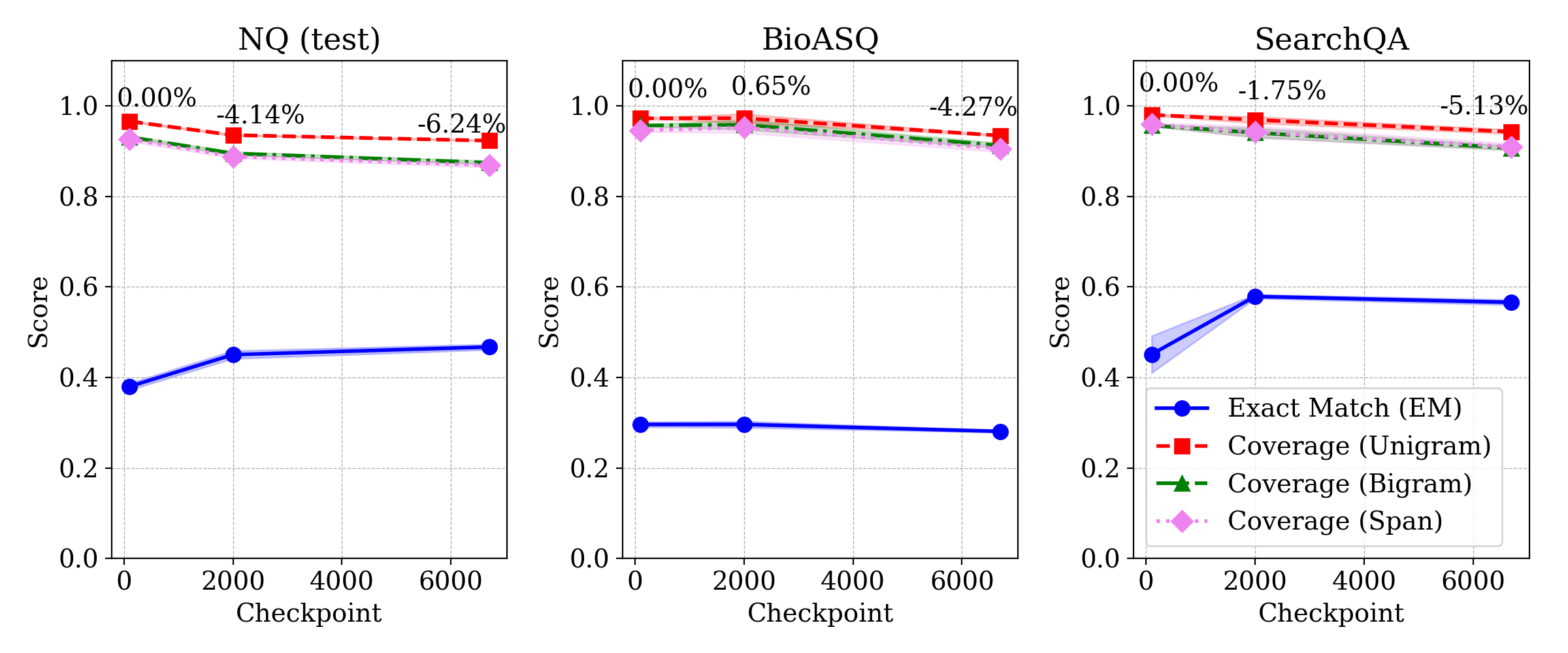}
 \caption{Individual faithfulness score and task performance with extractive QA datasets evaluated with three distinct model checkpoints of LLaMA-7B fine-tuned on context-dependent datasets. The middle checkpoint is the one with lowest in-training evaluation loss.}
 \label{fig:paper_quandary_stage_1}
\end{figure*}

\begin{figure*}[t]
 \centering
 \includegraphics[width=1.0\linewidth]{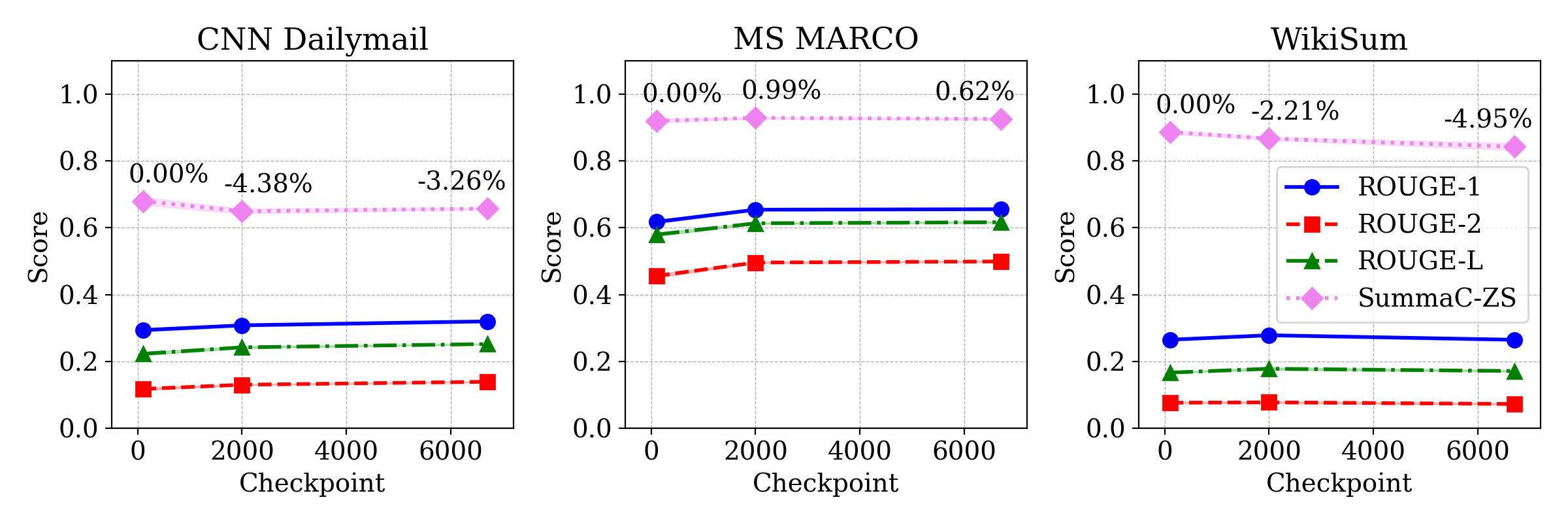}
 \caption{Individual faithfulness score and task performance with abstractive QA and summarization datasets across evaluated with three distinct model checkpoints of LLaMA-7B fine-tuned on context-dependent datasets. The middle checkpoint is the one with lowest in-training evaluation loss.}
 \label{fig:paper_abstract_stage_1}
\end{figure*}

\Figref{fig:paper_quandary_stage_1} shows faithfulness score and task performance across related datasets when we fine-tune LLaMA-7B on context-dependent datasets. One suprising finding is that throughout the fine-tuning process, the faithfulness scores on extractive QA datasets gradually decrease, while task performance scores gradually increase. On the other hand, this trend is not salient for abstractive tasks as shown in \Figref{fig:paper_abstract_stage_1}.

\subsection{Examples of Actual Instructions}\label{app:exampele_instructions}

From \Tabref{tab:nq_example} to \Tabref{tab:alpaca_example}, we show actual model inputs from each testing datasets.

\subsection{Qualitative Examples}\label{app:example_generation}

\Tabref{tab:first_pip_example} and \Tabref{tab:second_pip_example} show two qualitative examples of actual model generations from our experiments.

\begin{figure*}[t]
 \centering
 \includegraphics[width=1.0\linewidth]{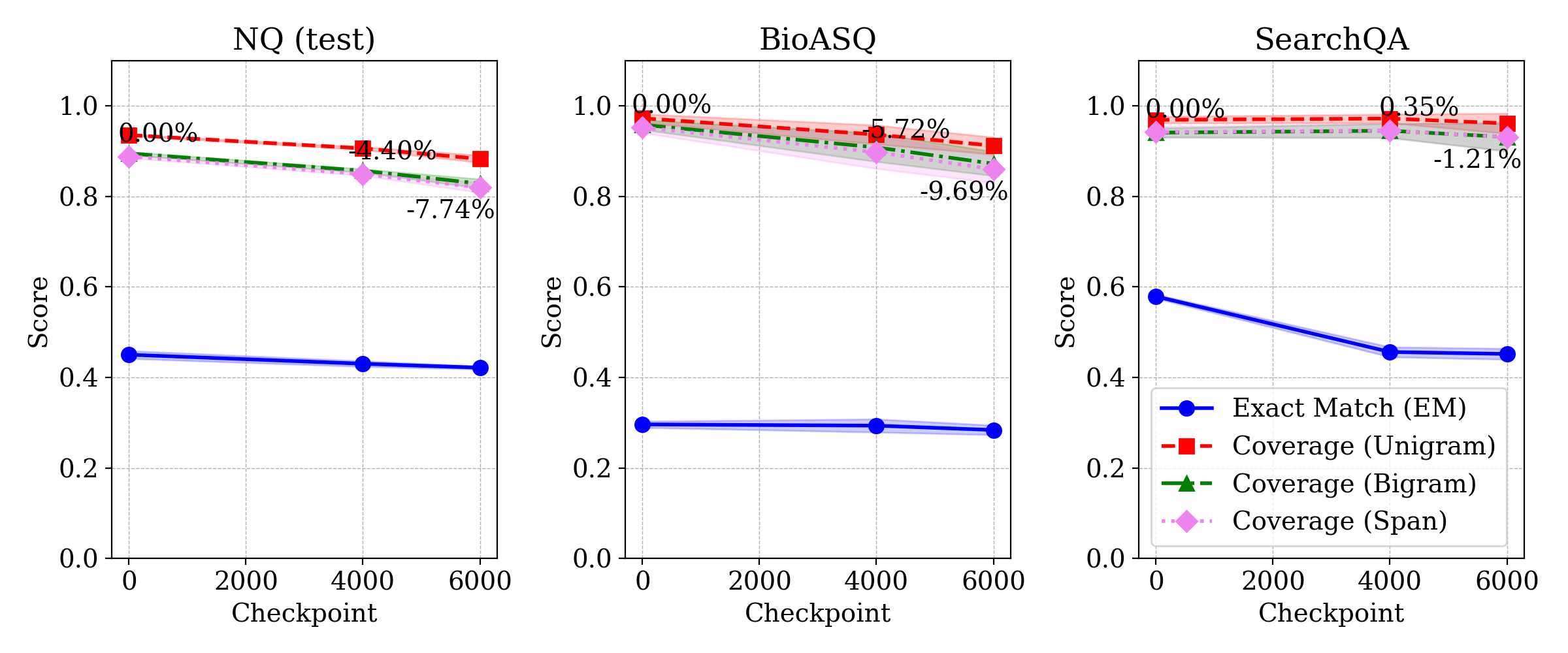}
 \caption{Individual faithfulness score and task performance with extractive QA datasets across evaluated with three distinct model checkpoints through the instruction following fine-tuning. The middle checkpoint is the one with lowest in-training evaluation loss.}
 \label{fig:paper_qa_summ_stage_2}
\end{figure*}

\begin{figure*}[t]
 \centering
 \includegraphics[width=1.0\linewidth]{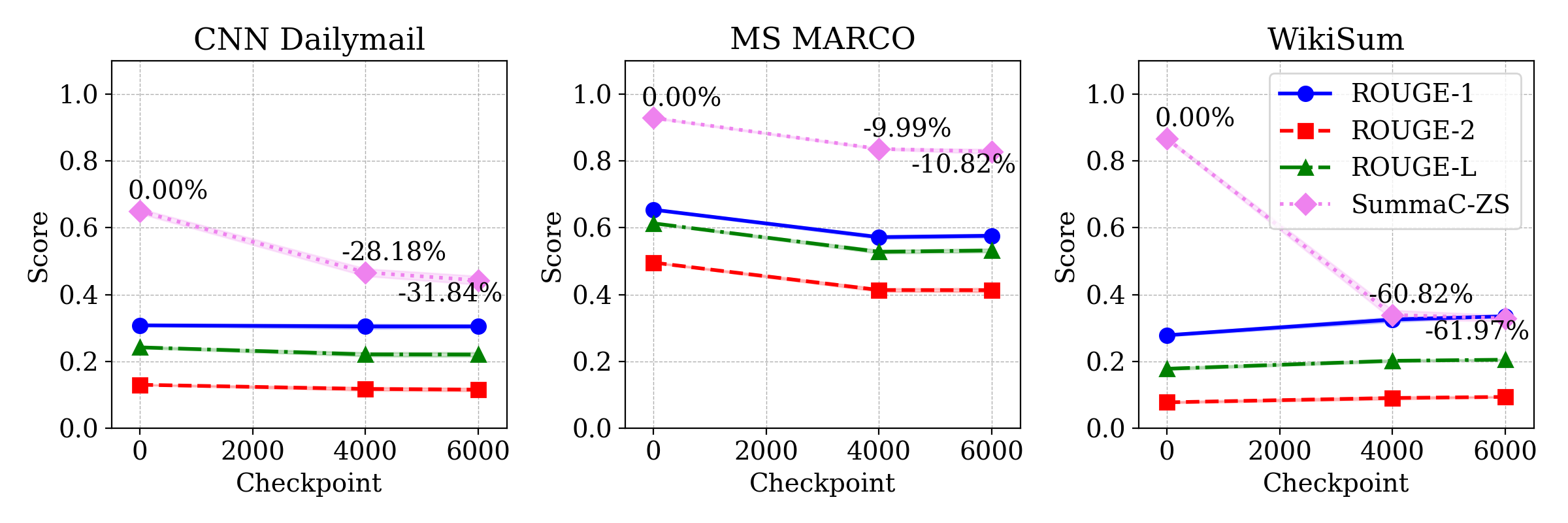}
 \caption{Individual faithfulness score and task performance with abstractive QA and summarization datasets across evaluated with three distinct model checkpoints through the instruction following fine-tuning. The middle checkpoint is the one with lowest in-training evaluation loss.}
 \label{fig:paper_abstract_stage_2}
\end{figure*}

\clearpage

\input{tables/first_pip_example}

\clearpage

\input{tables/second_pip_example}

\clearpage

\input{tables/nq_example}

\clearpage

\input{tables/cnn_dailymail_example}

\clearpage

\input{tables/ms_macro_example}

\clearpage

\input{tables/bioasq_example}

\clearpage

\input{tables/searchqa_example}

\clearpage

\input{tables/wikisum_example}

\clearpage

\input{tables/alpaca_example}

\clearpage

\end{document}

%% file: tables/first_pip_example.tex
\begin{table*}[ht]
    \centering
    \small
    \begin{tabular}{cp{14cm}}
\toprule
\multicolumn{2}{c}{\textbf{Testing Example from WikiSum Dataset}}\\
\midrule
Passage & Take your pencil and on the top of the page, about two inches from the right start drawing a line to 2 inches (5.1cm) below the right top corner. The line should loop down and have points. (see the picture) Draw straight lines from the points in your first line to the corner.  Make lines parallel to your fist line going all the way up. You should have 5 or 6 lines. Get a paper and make a cross on it, try to make both lines a similar length (using a ruler will help)  Draw diagonal lines through the centre, dividing the paper up from 4 to 8 sections. Make sure they are smaller than the cross you made before. Start connecting the lines with inverted arcs, this is an arc ), from the inside out.  Once you have reached the end of the web, elongate the diagonal lines, (this will make it look like it has supports).  Draw a spider by making a fuzzy ball, then drawing legs (eight of them) on your web. Or see the spider drawing tip. Finished.  Draw a circle and draw a cross section which also extends outside the circle.  Draw two diagonal lines at the midpoint of the cross-sections which form an X-mark.  Draw squares which descend in size as it nears the centre point. Draw the corners or vertices of the square along the diagonal lines. Draw diamond shapes descending in size as it nears the centre point. Draw the vertices along the lines of the cross-section. Draw curves to connect the lines \u2013 from the squares to the diamonds, much like forming bridges.  Trace with a pen and erase unnecessary lines. You may add drawings for spiders. Color to your liking! \\ 

\midrule
\multicolumn{2}{c}{\textbf{Golden Answer}}\\
\midrule

Response & To draw a spider web, start by drawing a cross with 2 lines of equal length. Next, draw diagonal lines through the center to divide your web into 4 to 8 equal sections. Then, connect the lines with inverted arcs, or curved lines, from the inside out. When you reach the end of the web, elongate the diagonal lines to make the web look like it's hanging from something, like a tree or shrub. Finally, draw a spider in your web by making a fuzzy ball with 8 legs. \\ 

\midrule
\multicolumn{2}{c}{\textbf{Response from LLaMA-7B Fine-tuned on Context-Dependent Datasets}}\\
\midrule

Response & Draw a cross on a piece of paper. Draw a spider by making a fuzzy ball, then drawing legs (eight of them) on your web. \\ 

\midrule
\multicolumn{2}{c}{\textbf{Response from LLaMA-7B Further Fine-tuned on Instruction Following Datasets}}\\
\midrule

Response & This is a simple spider web drawing tutorial. It is a simple and easy to follow tutorial. It is a great way to practice your drawing skills. \\ 

\bottomrule

\end{tabular}
    \caption{WikiSum testing example from the \textbf{first} pipeline of our two-stage fine-tuning paradigm where we take LLaMA-7B and fine-tune it on context-dependent datasets first, and then further fine-tune it with human alignment datasets.}

    \label{tab:first_pip_example}
\end{table*}

%% file: tables/second_pip_example.tex
\begin{table*}[ht]
    \centering
    \small
    \begin{tabular}{cp{14cm}}
\toprule
\multicolumn{2}{c}{\textbf{Testing Example from the Alpaca-15K Dataset}}\\
\midrule
Instruction & Construct a sentence that adequately describes a baby duck. \\ 

\midrule
\multicolumn{2}{c}{\textbf{Vicuna-7B Zero-Shot}}\\
\midrule

Response & A baby duck is a fluffy, downy creature with a round body and webbed feet, often seen quacking and waddling around with its mother. \\ 

\midrule
\multicolumn{2}{c}{\textbf{Examples from Vicuna-7B w/ Context Dependent Fine-tuning}}\\
\midrule

Response & A baby duck is a young duck that has not yet developed its full adult plumage. \\ 

\midrule
\multicolumn{2}{c}{\textbf{Examples from Vicuna-7B w/ MTL Fine-tuning}}\\
\midrule

Response & A baby duck is a small, fluffy, and adorable bird that is often seen waddling around with its parents. \\ 

\midrule
\multicolumn{2}{c}{\textbf{Examples from Vicuna-7B w/ \OurNameShort{}-S}}\\
\midrule

Response & A baby duck is a small, fluffy, and adorable bird that is often seen swimming in ponds or lakes. \\ 

\bottomrule

\end{tabular}
    \caption{Qualitative examples from the \textbf{second} pipeline of our two-stage fine-tuning paradigm where we take Vicuna-7B and fine-tune it on context-dependent datasets with various fine-tuning methods.}
    \label{tab:second_pip_example}
\end{table*}

%% file: tables/nq_example.tex
\begin{table*}[ht]
    \centering
    \small
    \noindent\fbox{%
    \begin{minipage}{2.0\columnwidth} 
\tt 
Below is an instruction that describes a task, paired with an input that provides further context. Write a response that appropriately completes the request.
\\

\#\#\# Instruction: \\
Answer to the question by extracting a specific text span from the given passages. Do not include new information beyond the given passages.
\\

\#\#\# Input:\\
Question: a bond that the issuer has the right to pay off before its maturity date
\\

Passage: does well. Companies also reserve the right to call their bonds, which mean they can call it sooner than the maturity date. Often there is a clause in the contract that allows this; for example, if a bond issuer wishes to rebuy a 30-year bond at the 25th year, they must pay a premium. If a bond is called, it means that less interest is paid out. Failure to pay a bond effectively means bankruptcy. Bondholders who have not received their interest can throw an offending company into bankruptcy, or seize its assets if that is stipulated in the contract.\\
Passage: Callable bond A callable bond (also called redeemable bond) is a type of bond (debt security) that allows the issuer of the bond to retain the privilege of redeeming the bond at some point before the bond reaches its date of maturity. In other words, on the call date(s), the issuer has the right, but not the obligation, to buy back the bonds from the bond holders at a defined call price. Technically speaking, the bonds are not really bought and held by the issuer but are instead cancelled immediately. The call price will usually exceed the par or issue\\
Passage: options embedded. Callable bond A callable bond (also called redeemable bond) is a type of bond (debt security) that allows the issuer of the bond to retain the privilege of redeeming the bond at some point before the bond reaches its date of maturity. In other words, on the call date(s), the issuer has the right, but not the obligation, to buy back the bonds from the bond holders at a defined call price. Technically speaking, the bonds are not really bought and held by the issuer but are instead cancelled immediately. The call price will usually exceed the par\\
Passage: the amount on which the issuer pays interest, and which, most commonly, has to be repaid at the end of the term. Some structured bonds can have a redemption amount which is different from the face amount and can be linked to the performance of particular assets. The issuer has to repay the nominal amount on the maturity date. As long as all due payments have been made, the issuer has no further obligations to the bond holders after the maturity date. The length of time until the maturity date is often referred to as the term or tenure or\\
Passage: Bond (finance) In finance, a bond is an instrument of indebtedness of the bond issuer to the holders. The most common types of bonds include municipal bonds and corporate bonds. The bond is a debt security, under which the issuer owes the holders a debt and (depending on the terms of the bond) is obliged to pay them interest (the coupon) or to repay the principal at a later date, termed the maturity date. Interest is usually payable at fixed intervals (semiannual, annual, sometimes monthly). Very often the bond is negotiable, that is, the ownership of the instrument can be
\\

\#\#\# Response:\\
\textbf{\underline{Model Generated Answer Goes Here}}
    \end{minipage}
}
    \caption{An example of NQ dataset with our instruction template for LLaMA-7B. Vicuna-7B's template is slightly different as shown in \Tabref{tab:vicuna_instruction_template}.}
    \label{tab:nq_example}
\end{table*}

%% file: tables/cnn_dailymail_example.tex
\begin{table*}[ht]
    \centering
    \small
    \noindent\fbox{%
    \begin{minipage}{2.0\columnwidth} 
\tt 

Below is an instruction that describes a task, paired with an input that provides further context. Write a response that appropriately completes the request.
\\

\#\#\# Instruction:\\
Summarize the text in a few sentences. Using original phrases or paraphrasing them if necessary. Do not include new information beyond the given passages.\\

\#\#\# Input:\\
(Billboard)Considering the Academy of Country Music Awards celebrated its 50th anniversary on Sunday night at the Dallas Cowboys stadium, it was bound to be bigger than any previous year's ACMs. Plus, as hosts Blake Shelton and Luke Bryan were quick to point out, everything is bigger in Texas. Billboard: 2015 ACM Awards: See All the Photos . But bigger isn't always better. Here's our breakdown of the 10 best and 5 worst moments at the 2015 ACMs. The Best . Eric Church \& Keith Urban provide an opening wallop . With a full stadium and millions of home viewers watching, two of country's leading men successfully lit the fuse for the 50th annual ACM Awards with a slick joint kickoff performance. Eric Church's \"Pledge Allegiance To The Hag\" was a fine throwback, but Keith Urban's powerhouse \"Raise 'Em Up\" lived up to its title -- and then some. Setting a Guinness World Record . Not only was this the ACMs biggest audience ever, but the 2015 ACM Awards brought in the biggest audience for a live TV awards show ever. Now that's how you celebrate half a century. Reba McEntire demonstrates how it's done . During a night that found a couple performers sounding a little weak in the vocal department, Reba McEntire showed the entire Cowboys stadium -- and many younger singers who don't have half her energy -- how it's done. Reba is eternal. Taylor Swift singing in the audience . Whether it was during Eric Church or Martina McBride, T-Swift was heating up the audience by singing and swaying along to country staples. Taylor might have gone pop for \"1989,\" but she was deep in the heart of Texas for this annual country extravaganza. Billboard: 2015 ACM Awards: And the Winners Are... Garth Brooks' All-American salute . During Brooks' performance of \"All-American Kid,\" the country giant welcomed a slew of U.S. military representatives into AT\&T Stadium, in an emotional moment that put the red, white and blue front and center. Brooks ended the performance by saluting the U.S. armed forces \"who are here and are all around the world for protecting our freedom,\" and also shouted out his home state of Oklahoma, who are grieving 20 years after the Oklahoma City bombing on Sunday. A truly classy moment from one of the best ever. Taylor Swift's mom makes an appearance . Sure, Andrea Swift's presentation of a special award for her daughter came with a fun origin tale about the creation of \"Love Story\" and some lovely words about the 25-year-old superstar. But the sight of the elder Swift -- just days after Taylor confirmed that her mother had been diagnosed with cancer -- walking to the podium was more than enough to yield one of the night's most poignant moments. Christina Aguilera joins Rascal Flatts . Aguilera is far from country, but with a voice as malleable as hers, she can pass for a song or two. After singing a bit of the tune she did while guesting on Nashville, Xtina joined longtime hitmakers Rascal Flatts for \"Riot\" from their recent album \"Rewind.\" Aguilera and Gary LeVox trading vocals was the rare unexpected artist pairing that actually worked. Miranda Lambert domination . In addition to kicking ass during her \"Mama's Broken Heart\"/\"Little Red Wagon\" medley, Lambert justly owned the night when it came to awards. If there's one thing the country community loves more than Miranda Lambert, it's giving Miranda Lambert awards. Billboard: Watch Little Big Town Bring Provocative 'Girl Crush' to ACMs . Little Big Town don't back down . Despite some mild controversy over their song \"Girl Crush,\" Little Big Town brought the poignant ballad to the awards show, giving the ACMs one of its more melancholy moments. Jason Aldean comes on strong . For a night featuring some shaky vocals, Aldean brought his silky yet powerful country croon to the ACMs during a massive medley. It's hard to see an audience get weak in the knees when you're watching at home, but it's fair to assume that's what happened during his performance. The Worst . Tony Romo . The Dallas Cowboys QB was understandably a little stiff on the mic (athletes usually aren't the most charismatic public speakers) but the whole gag with Shelton asking Romo to toss Bryan a pass went on waaaaay too long. On the plus side, Bryan caught the pass. On the other hand, there was a tired play on words about balls. The length . Three hours for the 50th ACMs? Sure, why not. Three and a half hours? That's pushing it. They could have shaved off the last half hour by cutting a couple of the performers who only sang half a song, and shortened a few of the massive commercial breaks. All of the Milestone Awards . Taylor Swift was given an extended honor at this year's ACM Awards, but some of the Milestone Awards -- especially those given to Reba McEntire, Kenny Chesney and George Strait -- seemed rushed for the country giants they were saluting. It's understandable since 2015 is the 50th anniversary of the ACMs, but sometimes, less (recipients) is more. Steven Tyler's facial hair . See link. Twitter calling out Taylor Swift . Plenty of country fans went after T-Swizzle on Twitter, berating her for attending the ACMs after \"abandoning\" country music for pop. The truth is, Swift has just as many country classics under her belt as any other artist in her age range. She might have moved to pop, but don't underplay her importance to the genre that birthed her. Billboard. All Rights Reserved.\\

\#\#\# Response:\\
\textbf{\underline{Model Generated Answer Goes Here}}
    \end{minipage}
}
    \caption{An example of CNN DailyMail dataset with our instruction template for LLaMA-7B. Vicuna-7B's template is slightly different as shown in \Tabref{tab:vicuna_instruction_template}.}
    \label{tab:cnn_example}
\end{table*}

%% file: tables/ms_macro_example.tex
\begin{table*}[ht]
    \centering
    \small
    \noindent\fbox{%
    \begin{minipage}{2.0\columnwidth} 
\tt 

Below is an instruction that describes a task, paired with an input that provides further context. Write a response that appropriately completes the request.\\

\#\#\# Instruction:\\
Answer the question with well-formed sentences. Paraphrase the context in the passages if necessary. Do not include new information beyond the given passages.\\

\#\#\# Input:\\
Question: 2015 college half term holiday dates\\

Passage: UK School Half Term Dates 2015, 2016 and 2017. Research the latest school holidays and term dates for England, Scotland, Wales and Ireland. Welcome to Half Term Dates. The place to view School Holidays for the UK, Ireland, France and Australia. We list the latest published half term times on one handy website.\\
Passage: UK School Half Term Dates 2015, 2016 and 2017. Research the latest school holidays and term dates for England, Scotland, Wales and Ireland. Welcome to Half Term Dates. The place to view School Holidays for the UK, Ireland, France and Australia.\\
Passage: UK School Half Term Dates 2015, 2016 and 2017. Research the latest school holidays and term dates for England, Scotland, Wales and Ireland.\\
Passage: Find your child\u2019s school term, half term and holiday dates on your local council 2019s website.\\
Passage: School term dates-guide. These are the school term and holiday dates for schools in Lambeth. The dates shown do not take account of the five professional development days when schools are closed to pupils, or any other changes.\\
Passage: Find your child 2019s school term, half term and holiday dates on your local council 2019s website. School term and holiday dates vary across the UK.\\
Passage: Irish School Holidays, Mid Term Dates. Research the official half term breaks and school holidays in Ireland. The Irish authorities do not enforce the same holidays as the UK. Noticeable differences being the lack of early June half term and a 9 week long summer holiday. Commencing end of June for most Irish schools.\\
Passage: School term dates-guide. These are the school term and holiday dates for schools in Lambeth. The dates shown do not take account of the five professional development days when schools are closed to pupils, or any other changes. Please check with your child's school for more detailed information.\\
Passage: School term dates and holidays 2014/15. Here are the school term dates and holidays for Sandwell's primary, secondary and special schools in 2014/15. Some academies and voluntary aided church schools may not follow this schedule. You are advised to check with these schools directly.\\
Passage: Holiday Dates for college students*: 1  Autumn Half Term: 26 October 2015 to 30 October 2015 (teaching re-starts from 3/4 November 2015 due to staff P\&D days 2013 check with your tutor). 2  Winter Break: 18 December 2015 to 5 January 2016 (teaching re-starts from 5/6 January 2016 due to staff P\&D days 2013 check with your tutor).\\

\#\#\# Response:\\
\textbf{\underline{Model Generated Answer Goes Here}}
    \end{minipage}
}
    \caption{An example of MS MARCO dataset with our instruction template for LLaMA-7B. Vicuna-7B's template is slightly different as shown in \Tabref{tab:vicuna_instruction_template}.}
    \label{tab:ms_macro_example}
\end{table*}

%% file: tables/bioasq_example.tex
\begin{table*}[ht]
    \centering
    \small
    \noindent\fbox{%
    \begin{minipage}{2.0\columnwidth} 
\tt 

Below is an instruction that describes a task, paired with an input that provides further context. Write a response that appropriately completes the request.\\

\#\#\# Instruction:\\
Answer to the question by extracting a specific text span from the given passages. Do not include new information beyond the given passages.\\

\#\#\# Input:\\
Question: abnormalities in which chromosomes were linked to the moyamoya disease?\\

Passage: moyamoya disease. We postulate that a protein encoded on chromosome 21 may be related to the pathogenesis of moyamoya disease. Although the neuronal substrate is abnormal in Down syndrome patients, recovery from hemiplegic stroke in patients with MM-DS is comparable to recovery in patients with primary moyamoya.\\
Passage: Moyamoya disease (MIM 252350) is characterized by stenosis or occlusion of the terminal portions of the bilateral internal carotid arteries and by abnormal vascular networks at the base of the brain. There is a high incidence of moyamoya disease in Asia, especially in Japan. Multifactorial inheritance is estimated with lambda(s)>40. Previous linkage studies have indicated that susceptibility loci for the disease are located on chromosomes 3p, 6q, and 17q. In the present study, we searched for loci linked to the disease in 12 Japanese families using 428 microsatellite markers and found significant evidence for linkage to 8q23 [maximum LOD score\\
Passage: We reported an autopsy case of Down's syndrome with moyamoya syndrome. A 30-year-old male with Down's syndrome suffered from a cerebral infarction and died of brain herniation. Cerebral angiography showed vascular abnormalities that were the same as moyamoya disease. Pathological findings revealed multiple stenosis of main trunk of the cerebral arteries. Pathologically, the stenosed vessels showed eccentric intimal thickness with cholesterin deposit, unlike moyamoya disease. There are only two previous reports of autopsied cases of Down's syndrome with moyamoya syndrome. We postulate that a protein encoded on chromosome 21 may be related to the pathogenesis of Down's syndrome with moyamoya\\
Passage: other). The karyotype was normal. No mutation in the RFN213 gene was found, and none of the HLA types linked to moyamoya disease or described in similar familial cases were identified. By describing these multisystemic associations, polycystic kidney disease for the second time, and intestinal malformation for the first time in the literature, our report expands the phenotypic variability of moyamoya syndrome. The coexistence of disparate malformations among close relatives suggests an underlying common genetic background predisposing to structural or physiological abnormalities in different tissues and organs.\\
Passage: OBJECTIVE: We report a detailed description of a family affected by a hereditary multisystem disorder associated with moyamoya syndrome.METHODS: In this family case report, we evaluated 9 members of the same family originating from Algeria. Investigations included neuroimaging, cardiologic and ophthalmologic evaluation, hormonal testing, hemoglobin electrophoresis, chromosomal karyotyping, muscle biopsy for morphology, immunohistochemistry and enzyme assays, mtDNA mutation screening, and haplotype analysis of 2 loci previously linked to moyamoya, on chromosomes 10 (ACTA2) and 17.RESULTS: Five males related through a maternal lineage were affected, suggesting an X-linked inheritance. Four of them had symptomatic moyamoya syndrome with an onset of acute\\

\#\#\# Response:\\
\textbf{\underline{Model Generated Answer Goes Here}}
    \end{minipage}
}
    \caption{An example of BioASQ dataset with our instruction template for LLaMA-7B. Vicuna-7B's template is slightly different as shown in \Tabref{tab:vicuna_instruction_template}.}
    \label{tab:bioasq_example}
\end{table*}

%% file: tables/searchqa_example.tex
\begin{table*}[ht]
    \centering
    \small
    \noindent\fbox{%
    \begin{minipage}{2.0\columnwidth} 
\tt 
Below is an instruction that describes a task, paired with an input that provides further context. Write a response that appropriately completes the request.\\

\#\#\# Instruction:\\
Answer to the question by extracting a specific text span from the given passages. Do not include new information beyond the given passages.\\

\#\#\# Input:\\
Question: \"30 days in the hole\" if you can't name this old peter frampton band whose name refers to a forced apology\\
Passage: CLASSIC YUMMY ROCKERS | \"30 Days in the Hole\" if you can't name this old \\
Peter Frampton band whose name refers to a forced apology | Humble Pie. right:.\\
Passage: 30 Days in the Hole by Humble Pie song meaning, lyric interpretation, video and \\
... Steve Marriot had said that Peter Frampton had heard the early stages of this ... \\
of mine with the 'Dead End Kids' from the movie of the same name, \"Dead End\", \\
... \"A Dirty room\": one full of evidence that can get you busted; especially if you...\\
Passage: \"30 Days in the Hole\" is the seventh single by English rock group Humble Pie, \\
from the band's ... The song refers to Borstal - \"some seeds and dust, and you got \\
Borstal\"- referring to Borstal Prison and its borstal ilk ... Humble Pie's Greatest \\
Hits \{Featuring Peter Frampton \& Steve Marriott]; Best of Humble Pie; Classics \\Volume...\\
Passage: Apr 1, 2012 ... Perhaps it was my new loose and mellow attitude, perhaps I had ... As I wrote \\
earlier this month, it is now illegal to sell an old piano with ... Needless to say this \\
would devastate the antique industry and force a ...... 30 Days In The Hole ... This \\
song of Humble Pie was recorded after Frampton left the band...\\
Passage: The Irish town of Kerry lends its name to this colorful breed of pooch | A Kerry \\blue .... Peter Frampton band whose name refers to a forced apology | Humble \\
Pie.\\

\#\#\# Response:\\
\textbf{\underline{Model Generated Answer Goes Here}}
    \end{minipage}
}
    \caption{An example of SearchQA dataset with our instruction template for LLaMA-7B. Vicuna-7B's template is slightly different as shown in \Tabref{tab:vicuna_instruction_template}.}
    \label{tab:search_qa_example}
\end{table*}

%% file: tables/wikisum_example.tex
\begin{table*}[ht]
    \centering
    \small
    \noindent\fbox{%
    \begin{minipage}{2.0\columnwidth} 
\tt 

Below is an instruction that describes a task, paired with an input that provides further context. Write a response that appropriately completes the request.\\

\#\#\# Instruction:\\
Summarize the text in a few sentences. Using original phrases or paraphrasing them if necessary. Do not include new information beyond the given passages.\\

\#\#\# Input:\\
. Condition score the horse Condition scoring is a process in which you assess the amount of fat the horse has on it. Condition scoring requires that you look at and feel the horse's body and assess the amount of fat it is carrying in specific areas. With condition scoring, you can evaluate whether the horse is in ideal condition or not. Once you look at and feel an area, you will write down your assessment in a chart made specifically for condition scoring. Condition scoring can take some instruction and practice, so you may want to consult with your veterinarian for some guidance on the procedure. A horse needs some fat to get it through the winter but it shouldn't have so much fat on it that it becomes obese. Use a weight tape. A weight tape is a tool used to approximate a horse's weight. It is a measuring tape that is wrapped around a horse's back and chest and the measurement markings are in pounds or kilograms. Using a weight tape will not give you a completely accurate measurement. It is only an estimate. It is best used for assessing change over time. Have the horse weighed. If you are bringing your horse to a veterinary clinic or a center that has a horse scale, then you can have the horse actually weighed. However, this is not usually available to horse owners on a regular basis. Using a scale is the most accurate way to weigh a horse. Measure the horse's weight regularly. In order to get an accurate understanding of changes to the horse's weight over time, you will need to measure it on a regular basis. If you are very concerned about a horse's weight, this can be every week. If you think the horse is maintaining weight fine, then every couple of weeks should be fine. Use the same type of weight measurement every time you measure the horse. This is the only way to really assess changes in weight over time. Take the horse's winter coat into consideration. When a horse is in the cold during the winter, it can grow a thick winter coat. This can interfere with weight tape measurements and condition scoring, which is why it is important to put your hands on your horse as you take these measurements. This will enable you to feel for fat pockets and ribs beneath the horse's wooly coat. With this coat change in mind, try to be consistent with the pressure and placement of weight tape and how you feel the horse's body when condition scoring. If you focus on consistency, any changes you document will usually signal a real trend in weight change. Be sure to remove a horse's blanket daily to assess weight gain or loss. Record your horse's weight over time. To track the horse's weight, you will need to know what its weight was before the winter. Then, you will need to keep a record of its weight throughout the winter. Make sure to write down each weight along with the date in a journal or notebook. This will be helpful to share with your horse's veterinarian later on. Be sure that you measure the horse's weight the same way every time. For instance, if you use a horse tape to measure the weight in the Fall, then use it to make subsequent measurements. Get professional help. If you are unsure if your horse's weight loss is of concern or you are unsure how to help your horse gain weight, you should get some professional advice. Talk to your veterinarian about what and how much you should be feeding your horse to fatten it up. Your veterinarian will also be able to recommend further bloodwork and diagnostics if they suspect that something else, such as parasites or disease, may be causing your horse's weight loss. Weight loss, if at all, should be gradual. Contact your veterinarian immediately if you notice extreme changes in your horse's weight over a short period of time. Check your horse's manure. Your horse's manure can provide some helpful clues about their eating and drinking habits. If you notice anything different, then call your veterinarian. Some instances where you would want to call your horse's vet include stool that is: Runny or wetter than usual, such as diarrhea. Drier than usual and/or less frequent, which may indicate constipation. This could be caused by not being able to access water because it is frozen. Check your horse's water often in cold weather to make sure it is not frozen. A different texture or color than is normal for your horse. Identify the signs of an emergency. If your horse has lost too much weight it can be an emergency that requires immediate veterinary care. Assess your horse's vital signs. Make sure that it is breathing regularly and that it has a strong pulse. A horse with irregular breathing or a weak pulse, or that is weak and unable to move, is in desperate need of emergency veterinary care. If any of its vital signs are not strong, you should call a veterinarian immediately. Help your horse regain weight. Horses should eat 2 - 2.5\% of their body weight in good quality bulky foods, such as grass or hay, during winter. For the average 1,000 pound horse, this would mean eating 25 pounds. Make sure that your horse is getting enough. Getting your horse to gain weight requires that you feed it more calories every day than you have been. Discuss making one of the following changes to your horse's diet with your veterinarian: Give the horse unrestricted access to hay around the clock. Change to a higher grade of hay. Add in a complete feed. Add 4 to 8 ounces of corn or cocosoya oil to a small amount of grain. Add beet pulp, sweet feed, or concentrates to your horse's diet. \\

\#\#\# Response:\\
\textbf{\underline{Model Generated Answer Goes Here}}
    \end{minipage}
}
    \caption{An example of WikiSum dataset with our instruction template for LLaMA-7B. Vicuna-7B's template is slightly different as shown in \Tabref{tab:vicuna_instruction_template}.}
    \label{tab:wikisum_example}
\end{table*}

%% file: tables/alpaca_example.tex
\begin{table*}[ht]
    \centering
    \small
    \noindent\fbox{%
    \begin{minipage}{2.0\columnwidth} 
\tt 
Below is an instruction that describes a task, paired with an input that provides further context. Write a response that appropriately completes the request.\\

\#\#\# Instruction:\\
Compile a list of ethical considerations for the development of AI systems.\\

\#\#\# Input:\\
\\

\#\#\# Response:\\
\textbf{\underline{Model Generated Answer Goes Here}}
    \end{minipage}
}
    \caption{An example of Alpaca-15K with our instruction template for LLaMA-7B. Vicuna-7B's template is slightly different as shown in \Tabref{tab:vicuna_instruction_template}.}
    \label{tab:alpaca_example}
\end{table*}